\documentclass{article}

\usepackage{times}
\usepackage{graphicx} 
\usepackage{caption}
\usepackage{subcaption}

\usepackage{natbib}

\usepackage{algorithm}
\usepackage[noend]{algorithmic}

\usepackage{hyperref}

\usepackage[accepted]{icml2016} 

\usepackage{url}

\usepackage{multirow,color,graphicx}
\usepackage{subfloat}
\usepackage{amsmath,amssymb,amsthm,xfrac,amsfonts}
\usepackage[titletoc]{appendix}
\usepackage{xspace}
\usepackage{savesym,verbatim}
\usepackage[titletoc]{appendix}
\usepackage{subscript}

\newtheorem{lemma}{Lemma}
\newtheorem{theorem}{Theorem}
\newtheorem{proposition}{Proposition}

\newtheorem{corollary}{Corollary}

\expandafter\def\expandafter\normalsize\expandafter{%
    \normalsize
    \setlength\abovedisplayskip{2.0pt}
    \setlength\belowdisplayskip{2.0pt}
    \setlength\abovedisplayshortskip{2pt}
    \setlength\belowdisplayshortskip{2pt}
}

\usepackage{enumitem}
\setitemize{noitemsep,topsep=1pt,parsep=1pt,partopsep=1pt, leftmargin=12pt}
\makeatletter

\newcommand{\Rmnum}[1]{\expandafter\@slowromancap\romannumeral #1@}
\makeatother
\usepackage{caption}

\usepackage{dsfont}
\usepackage{mathtools}

\newcommand{\problems}{K}
\newcommand{\solutionspace}{S}

\newcommand{\loss}{l}

\newcommand{\w}{\pmb{w}}
\newcommand{\wtilde}{\widetilde{\w}}

\newcommand{\wstar}{\pmb{u}}
\newcommand{\Q}{\pmb{Q}}
\newcommand{\gradient}{\pmb{g}}
\newcommand{\z}{z}

\newcommand{\counter}{\tau}

\newcommand{\maxgradient}{\gradient}
\newcommand{\maxdualitygap}{\delta^t}
\newcommand{\dualitygap}{\delta}

\newcommand{\calpha}{c_\alpha}
\newcommand{\cbeta}{c_\beta}

\newcommand{\OCP}{\textsc{OL}\xspace}
\newcommand{\IOCP}{\textsc{IOL}\xspace}
\newcommand{\COCP}{\textsc{CoOL}\xspace}
\newcommand{\AdaGrad}{\textsc{AdaGrad}\xspace}
\newcommand{\ALG}{\textsc{Alg}\xspace}
\newcommand{\aprxproj}{\textsc{AprxProj}\xspace}

\newcommand{\bad}{\mathds{1}_{\{(\neg \xi^{t-1}) \wedge (\xi^{t})\}}}
\newcommand{\good}{\mathds{1}_{\{(\xi^{t-1}) \vee (\neg \xi^{t})\}}}
\newcommand{\outside}{\mathds{1}_{\{\neg \xi^t\}}}
\newcommand{\inside}{\mathds{1}_{\{\xi^t\}}}

\newcommand{\insidet}[1]{\mathds{1}_{\{\xi^{#1}\}}}
\newcommand{\outsidet}[1]{\mathds{1}_{\{\neg \xi^{#1}\}}}

\newcommand{\argmin}{\mathop{\mathrm{argmin}}\limits} 
\newcommand{\norm}[1]{\left\lVert#1\right\rVert}

\DeclarePairedDelimiterX{\normnew}[1]{\big\lVert}{\big\rVert}{#1}

\newcommand{\names}[1]{\texttt{#1}}

\newcommand{\regret}{Regret}
\newcommand{\expectedvalue}[1]{\mathbf{E} \left[#1\right]}
\makeatletter											
\def\thm@space@setup{\thm@preskip=0.15\baselineskip		
\thm@postskip=0.15\baselineskip}							
\makeatother

\newcommand{\what}{\widehat{\w}}
\newcommand{\x}{\pmb{x}}
\newcommand{\zt}{{z^t}}

\newcommand{\da}{\Delta^t_1}
\newcommand{\db}{\Delta^t_2}

\newcommand{\regularizer}{R}
\newcommand{\divergence}{D}
\newcommand{\projection}[2]{\prod_{#1} \ifthenelse{\equal{#2}{}}{}{\left(#2\right)}}

\icmltitlerunning{Coordinated Online Learning}

\begin{document}

\twocolumn[
\icmltitle{Coordinated Online Learning\\ \textnormal{With Applications to Learning User Preferences}}


\icmlauthor{Christoph Hirnschall}{chirnsch@student.ethz.ch}
\icmlauthor{Adish Singla}{adish.singla@inf.ethz.ch}
\icmlauthor{Sebastian Tschiatschek}{sebastian.tschiatschek@inf.ethz.ch}
\icmlauthor{Andreas Krause}{krausea@ethz.ch}
\icmladdress{ETH Zurich, Switzerland}

\vskip 0.3in
]


\begin{abstract}
We study an online multi-task learning setting, in which instances of \emph{related} tasks arrive sequentially, and are handled by task-specific online learners. We consider an algorithmic framework to model the relationship of these tasks via a set of convex constraints. To exploit this relationship, we design a novel algorithm -- \COCP\ -- for coordinating the individual online learners: Our key idea is to coordinate their parameters via \emph{weighted projections} onto a convex set. By adjusting the rate and accuracy of the projection, the \COCP algorithm allows for a trade-off between the benefit of coordination and the required computation/communication. We derive regret bounds for our approach and analyze how they are influenced by these trade-off factors. We apply our results on the application of learning users' preferences on the \names{Airbnb} marketplace with the goal of incentivizing users to explore under-reviewed apartments. 
\end{abstract}
\section{Introduction}\label{sec.introduction}
Many real-world applications involve a number of different learning tasks. Very often, these individual tasks are \emph{related}, and by sharing information between these tasks, we can improve the performance of the overall learning process. For instance, wearable devices that provide personalized recommendations to users can improve their performance by leveraging the knowledge of data from other users. This idea forms the basis of multi-task learning \cite{caruana1998multitask}.

In this paper, we study multi-task learning in the framework of \emph{online regret minimization} (\emph{cf.} \cite{cesa2006prediction,shalev2011online}). We investigate the problem of online learning of $\problems$ related tasks (or classes/types of problems) jointly.
For each task $\z \in [\problems]$, we have a separate online learner $\OCP_z$ tackling the instances of this task. Task instances arrive in an arbitrary, possibly adversarial, order and each time $t$ corresponds to an instance of a task $\z^t$ which is received by learner $\OCP_{\z^t}$. Our goal is to coordinate the individual learners to exploit the relationship of the tasks and improve the  overall performance given by the sum of regrets over all $\problems$ learners.

\subsection{Motivating applications}\label{sec.introduction.app}
{\bfseries Personalized AI on wearable devices.}  Wearable devices such as \names{Apple Watch} or \names{Fitbit}, equipped with various sensors, aim to provide realtime predictions to the users, \emph{e.g.}\ for healthcare monitoring, based on their individual activity patterns~\cite{jin2015collaborating}. In this application, a task $\z$ corresponds to providing personalized predictions to a specific user, and the learner $\OCP_\z$ corresponds to an online learning algorithm implemented on her device. The relationship of the tasks could, for instance, help to enforce some smoothness in the predictions for users with similar demographics.

{\bfseries Hemimetrics encoding users' preferences.} Another motivating application, which forms the basis of the experiments in this paper (\emph{cf.} Section~\ref{sec.userstudy}), is to learn preferences of a user (or cohort of users) across different choices. These choices take the form of $n$ items available in a marketplace (\emph{e.g.} items could correspond to different apartments on \names{Airbnb}). 
Our goal is to learn the pairwise distances $D_{i,j}$ representing the private cost of a user for switching from her default choice of item $i$ to item $j$.
Knowledge about this type of preferences can be used in e-commerce applications for marketing or for maximizing social welfare, \emph{e.g.} by persuading users to change their decisions \cite{singla2016actively,kamenica2009bayesian,singla2015incentivizing}. 
The interaction with the users takes the form of a binary query, motivated by the \emph{posted-price model} in marketplaces  \cite{abernethy2015low,singla2013truthful},  where users are offered a take-it-or-leave-it offer price that they can accept or reject.
The goal is to learn these $\problems = n^2-n$ distances  while interacting with the users sequentially, where learning each such pairwise distance $D_{i,j}$ corresponds to one task.
These distances are often correlated in real-world applications, for instance, satisfying \emph{hemimetric} properties, \emph{i.e.} a relaxed form of a metric \cite{singla2016actively}.

In this paper, we develop an algorithmic framework to model such complex dependencies among online learners and \emph{efficiently} coordinate their learning process.

\subsection{Our Approach}
{\bfseries Relationship of the tasks.}  In multi-task learning, one of the key aspects is modeling the relationship of the tasks. A common approach is to consider a specific structure capturing this relationship and develop an appropriate algorithm exploiting this structure: examples of this approach include shared parameters among the tasks \cite{chapelle2010multi,jin2015collaborating}, shared support \cite{wang2016distributed}, and smoothness in the parameters \cite{zhou2013modeling}. In this paper, we develop an algorithmic framework to model this relationship via a set of convex constraints. Our approach captures some of the above-mentioned structures, and allows us to model more complex dependencies.

{\bfseries Coordination via weighted projection.}  
Given a generic convex set capturing task relatedness/structure, a natural question is how to design efficient algorithms to coordinate the learners to exploit this structure. For this setting, we present a principled way to coordinate learners: we show that coordination can be achieved via \emph{weighted projection} (with carefully chosen weights for each learner) of the current solution vectors of the learners onto the convex set defined by the structural constraints. 

{\bfseries Sporadic and approximate coordination.} 
For large scale applications (\emph{i.e.} large $\problems$), coordination at every step via projection could be computationally very expensive. Furthermore, when applying our framework in a distributed setting, it is often desirable to design a communication-efficient coordination protocol. In order to make our approach applicable in these settings, we employ two algorithmic ideas: sporadic and approximate coordinations (\emph{cf.} Section~\ref{sec.algorithm} for details). This allows us to speed up the algorithm by an order of magnitude while retaining the improvements obtained by coordination (\emph{cf.} Section~\ref{sec.experiments}). Furthermore, these two features have privacy-preserving properties that can be exploited in the distributed setting, if privacy or data leakage across the learners is a concern \cite{balcan2012distributed}: for instance by using ideas of differential privacy in online learning (\emph{cf.} \cite{jain2012differentially}) to control the accuracy of the projection.

%
%
%


%
%
%
%

\subsection{Main Contributions}
We develop a novel algorithm -- \COCP\ -- that employs the above-mentioned ideas to coordinate the individual online learners. We consider a standard adversarial online setting \cite{cesa2006prediction,shalev2011online} without any probabilistic assumptions on the loss functions or order of task instances. We derive regret bounds for our approach and provide insights into how the trade-off factors controlling the rate/accuracy influence these bounds. We perform extensive experiments for learning hemimetric structures encoding users' preferences to support the conclusions of our theoretical analysis. Furthermore, we collect data via a survey study on the \names{Airbnb} marketplace and  demonstrate the practical applicability of our approach through experiments on this dataset.
\section{Preliminaries}\label{sec.model}
We now formalize the problem addressed in this paper.
\subsection{The Model and Protocol}

{\bfseries Tasks and Learners.}
We consider a set of $\problems$ tasks (or types/classes of problems). For each task $\z \in [\problems]$, we have a separate online learner $\OCP_\z$ tackling the instances of this task. 
For each task $\z$, the online learner $\OCP_\z$ learns some model parameters denoted by a weight vector $\w_\z \in \solutionspace_\z \subseteq \mathbb{R}^{d_\z}$, where $\solutionspace_\z$ denotes the feasible solution space. For simplicity of notation and w.l.o.g.\ we assume that $d_\z  = d \ \forall z \in [\problems]$. Using the standard convex online learning framework~\cite{shalev2011online,zinkevich2003online}, we assume that $\solutionspace_\z$ is a convex, non-empty, and compact set:  
Let $\norm{\solutionspace_\z}$ denote the diameter of the solution space for task $\z$ (w.r.t the Euclidean norm)\footnote{Euclidean norm is used throughout, unless specified.}. We assume $\norm{\solutionspace_\z} \leq \norm{\solutionspace_{\textnormal{max}}}$ for some constant $\norm{\solutionspace_{\textnormal{max}}}$.

{\bfseries Online protocol.} We consider an online setting, where each round is indexed by time $t$. Each learner $\OCP_z$ maintains  a weight vector $\w^t_\z$.  At time $t$, the environment generates an instance of task $\z^t$, which is received and handled by online learner $\OCP_{\z^t}$. Hence, at time $t$, the learner $\OCP_{\z^t}$ extends its prediction $\w^t_{\z^t}$, suffers a loss $\loss^t(\w^t_{\z^t})$, and updates its parameters to obtain $\w^{t+1}_{\z^t}$. All other learners $\OCP_{\z} \ \forall \z \neq \z^t$ do not interact with the environment at this time, \emph{i.e.} they do not update their parameters  ($\w^{t+1}_{\z} = \w^t_{\z}$).  Again, using the convex online learning framework~\cite{shalev2011online,zinkevich2003online}, we consider the loss functions to be convex, \emph{i.e.}\ $\loss^t(\w^t_{\z^t})$ is convex w.r.t.\ the parameter $\w^t_{\z^t}$ for all $t$. We further consider gradient-based learners, and assume that learner $\OCP_{\z^t}$ at the end of round $t$ has access to the (sub-)gradient $\gradient^t_{\z^t}$ of the loss function $\loss^t$ computed at $\w^t_{\z^t}$.\footnote{This is a weaker requirement than the full-information model where we assume access to the function $\loss^t$ and discuss this point further in our experiments, \emph{cf.} Section~\ref{sec.userstudy}.} For a given task $\z$, we assume that the Euclidean norm of (sub-)gradients $\norm{\gradient^t_{\z^t}}$ is upper bounded by $\norm{\maxgradient_{z}}$ whenever $\z^t = \z$. Furthermore, $\norm{\maxgradient_{z}} \leq \norm{\maxgradient_{\textnormal{max}}}$ where $\norm{\maxgradient_{\textnormal{max}}}$ is a constant. We consider a standard adversarial online setting \cite{cesa2006prediction,shalev2011online} without any probabilistic assumptions on the loss functions.

{\bfseries Order of task instances.} We consider a general setting where the order of the task instances and the total number of instances of any given task is arbitrary. Furthermore, in our setting each time step is associated with one task instance only. This is strictly more general than the synchronized setting, in which all the task instances arrive in parallel, \emph{e.g.} as required in \cite{dekel2007online,lugosi2009online}. When implementing our algorithmc ideas for distributed optimization problems (\emph{cf.}\ \cite{wang2016distributed,dekel2012optimal,shamir2014distributed}), the coordinating algorithm (\emph{e.g.} implemented via the master node in a cluster) can control the schedule of the tasks, and our results directly apply in these more controlled settings as well.




\subsection{Relationship of the tasks} \label{subsec.relatedness}
We denote the joint solution space of the $\problems$ tasks as $\solutionspace =  \solutionspace_1 \times {\cdots} \times \solutionspace_\z  \times {\cdots} \times \solutionspace_\problems \subseteq \mathbb{R}^{d \cdot {\problems}}$. Let $\w^*_\z \in \solutionspace_\z$ denote a \emph{competing} weight vector for task $z$ against which we compare the regret of learner $\OCP_z$. For instance, $\w^*_\z$ could be the optimal weight vector for task $z$ in hindsight. We define a joint competing weight vector $\w^* \in \solutionspace \subseteq \mathbb{R}^{d \cdot {\problems}}$ as the concatenation of the task specific weight vectors, \emph{i.e.} 
\begin{align*}
\w^*=\big[(\w^*_1)' \ {\cdots} (\w^*_\z)' \ {\cdots} (\w^*_\problems)'\big]'
\end{align*}
where $(.)'$ denotes the transposition operator and $\w^*_\z \ \forall \z \in [\problems]$ are column vectors. Similarly, we define $\w^t \in \solutionspace$ to be the concatenation of the task specific weight vectors at time $t$, \emph{i.e.} 
\begin{align*}
\w^t=\big[(\w^t_1)' \ {\cdots} (\w^t_\z)' \ {\cdots} (\w^t_\problems)'\big]'.
\end{align*}
We model the relationship of the tasks by using the following structural information: The joint competing weight vector $\w^*$ against which the regret of all the learners is measured lies in a convex, non-empty, and closed set $\solutionspace^* \subseteq \solutionspace$ representing a restricted joint solution space, \emph{i.e.} $\w^* \in \solutionspace^*$. 
This set $\solutionspace^*$ can be interpreted as the prior knowledge available to the algorithm that restricts the joint weight vector $\w^*$ to $\solutionspace^*$ (\emph{e.g.} hemimetric structure over the pairwise distances, \emph{cf.} Section~\ref{sec.introduction}).
Note that we do not require $\w^t \in \solutionspace^*$ at any given time $t$. 
However, our approach can also be specialized to the setting with \emph{hard constraints} over the tasks' joint weight vectors (\emph{e.g.} considered in \cite{lugosi2009online}), that would require $\w^t \in \solutionspace^* \ \forall t$. For this setting, we can enforce $\w^t \in \solutionspace^* \ \forall t$ by coordinating at every time step $t$ by using $\xi^t = 1 \ \forall t = 1$, \emph{cf.} Algorithm~\ref{algo.COCP}.

Next, we present three examples illustrating the kind of task relationships captured by the above model.

{\bfseries Unrelated tasks.} $\solutionspace^* \equiv \solutionspace$ models the setting where the tasks/learners are unrelated/independent.

{\bfseries Shared parameters.} A commonly studied setting in the distributed  stochastic optimization is parameter sharing. To model this setting, $\solutionspace^*$ is given as 
\begin{align*}
\solutionspace^* = \{\w^* \in \solutionspace \ | \ \w^*_1 = \cdots = \w^*_\z  = \cdots = \w^*_\problems\}.
\end{align*}
Instead of sharing all the parameters, another common scenario in multi-task learning is to share a few parameters. For a given $d' \leq d$, sharing $d'$ parameters across the tasks can be modeled by as 
\begin{align*}
\solutionspace^* = \{\w^* \in \solutionspace \ | \ \w^*_1[1\!:\!d']  \cdots = \w^*_\z[1\!:\!d']  = \cdots \w^*_\problems[1\!:\!d']\}
\end{align*}
where $\w^*_\z[1\!:\!d']$ denotes the first $d'$ entries in $\w^*_\z$. 

{\bfseries Hemimetric structure.} As discussed in Section~\ref{sec.introduction}, for $n$ items available in the marketplace, the $\problems$ tasks can be  represented as $\z = (i,j) \ \forall i,j \in [n], i \neq j$, where $\problems = n^2 - n$.
For $d=1$, the convex set representing $r$-bounded hemimetrics is given by $\solutionspace^* =$
\begin{align*}
\{\w^* \in \solutionspace \ | \ \w^*_{i,j} \in [0,r], \w^*_{i,j} \leq \w^*_{i,k} + \w^*_{k,j} \ \forall i, j, k \in [n]\}
\end{align*}
This structure is useful to model users' preferences \cite{singla2016actively} and considered in the experiments in Section~\ref{sec.experiments} and Section~\ref{sec.userstudy}. 

Overall, the framework of modeling the relationships via a convex set representing structural constraints is very general and can capture many complex real-world dependencies among the tasks/learners.

%
%

\subsection{Objective}
 
The focus of this paper is to develop an algorithm that plays the role of central coordinator. We measure the overall performance of the algorithm by the sum of cumulative losses of the $\problems$ individual learners. As is common in the online regret minimization framework (\emph{cf.} \cite{cesa2006prediction,shalev2011online}), we use regret, \emph{i.e.} loss of the algorithm w.r.t.\ the loss of a fixed competing  weight vector in hindsight, as performance measure. Considering a time horizon of $T$, the regret of the coordinating algorithm $\ALG$ against \emph{any} competing weight vector $\w^* \in \solutionspace^*$ is
\begin{align}
\regret_{\ALG}(T,\w^*) =  \sum_{t = 1}^{T} \Big(\loss^t(\w^t_{\z^t}) - \loss^t(\w^*_{\z^t})\Big). \label{eq.regretTotal1}
\end{align}
This can equivalently be written as sum of the regrets of $\problems$ individual learners. 
The objective is to develop an algorithm with low regret.

\section{Methodology}\label{sec.algorithm}
In this section, we develop our main algorithm \COCP. We begin with the specification of the individual learners $\OCP_z$, and also present a baseline algorithm \IOCP without coordination among learners.
\subsection{Specification of Online Learner $\OCP_z$}
In this paper, we consider the popular algorithmic framework of online convex programming (OCP) \cite{zinkevich2003online} for the individual learners $\OCP_\z \ \forall \z \in [\problems]$, each learning the corresponding task $\z$ separately. The OCP algorithm is a gradient-descent style algorithm, similar to the online-mirror descent family of algorithms (\emph{cf.} \cite{shalev2011online}), except that it performs a projection after every gradient step to maintain feasibility of the current weight vector.  
In order to formally describe the algorithm, let us consider a single-task setting where $\z^t = \z \ \forall t \in [T]$. The learner $\OCP_\z$ implementing the OCP algorithm uses the learning rate $\eta^t_\z$ and  updates the weight vector $\w^t_\z$ in each round as follows:
\begin{align}
\wtilde^{t+1}_\z = \w^t_\z - \eta^t_\z \gradient^t_\z; \ \ \w^{t+1}_\z = \argmin_{\w \in \solutionspace_\z}  \normnew{\w - \wtilde^{t+1}_\z}_2 \label{eq.OCP-update}
\end{align}

In this single-task setting, the regret against any competing weight vector $\w_\z \in  \solutionspace_\z$ after $T$ rounds is
\begin{align}
\regret_{\OCP_\z}(T, \w_\z) = \sum_{t=1}^{T} \Big(\loss^t(\w^t_\z) - \loss^t(\w_\z)\Big). \label{eq.regretOCP}
\end{align}
Theorem~\ref{theorem:OCP} below bounds the regret of learner $\OCP_\z$. 
\begin{theorem} [From  \cite{zinkevich2003online}] \label{theorem:OCP}
Consider the single-task setting where $\z^t = \z \ \forall t \in [T]$. For learning rate $\eta^t_\z = \frac{\eta}{\sqrt{t}}$ where $\eta = \frac{\norm{\solutionspace_{max}}}{\norm{\maxgradient_{max}}}$, the regret of the learner $\OCP_\z$ implementing the OCP algorithm is bounded as
\begin{align*}
\regret_{\OCP_\z}(T) \leq \frac{3}{2} \sqrt{T} \norm{\solutionspace_{max}} \norm{\maxgradient_{max}}.
\end{align*}
\end{theorem}

\begin{algorithm}[!t]
	\caption{Central coordinator \COCP}\label{algo.COCP}
\begin{algorithmic}[1]
	\STATE{	{\bfseries Input:}  
	    \begin{itemize}
    			\item Coordination steps: $(\xi^t)_{t \in [T]} \textnormal{ where } \xi^t \in \{0,1\}$ \\
    			\item Coordination accuracy:  $(\dualitygap^t)_{t \in [T]} \textnormal{ where } \dualitygap^t \geq 0$ \\
    		\end{itemize}
    }
	\FOR{$t = 1, 2, \ldots, T$}  
		\IF{$\xi^t = 1$}
			\STATE{ $\forall \z \in [\problems]\!:$  \texttt{RECEIVE} $\w_\z, \counter^t_{\z}$ \texttt{FROM} $\OCP_z$ \label{alg1.receive} }
			\STATE{ Define $\widetilde{\w} =\big[(\w_1)' \ {\cdots} (\w_\z)' \ {\cdots} (\w_\problems)'\big]'$ \label{alg1.wold}}
			\STATE{ Define $\Q^t$ as per Equation~\eqref{eq.qmatrix} }\label{alg1.Q}
			\STATE{ Compute $\w^{t+1} = \aprxproj(\widetilde{\w}, \dualitygap^t, \Q^t)$ }\label{alg1.wnew}
			\STATE{ $\forall \z \in [\problems]\!:$ \texttt{SHARE} $\w^{t+1}_\z$ \texttt{WITH} $\OCP_z$ }\label{alg1.share}
		\ENDIF
	\ENDFOR 
\end{algorithmic}
\end{algorithm}

\begin{algorithm}[!t]
	\caption{Function \aprxproj}\label{algo.aprxproj}
\begin{algorithmic}[1]
\STATE{	{\bfseries Input:} $\widetilde{\w}, \dualitygap^t, \Q^t$ }
\STATE{	Define $f^t(\w) = (\w - \wtilde)' \Q^t (\w - \wtilde)$ for $\w \in \solutionspace$ }
\STATE{	Choose $\w^{t+1} \in \{\w \in \solutionspace^* | f^t(\w) - \min_{\w' \in \solutionspace^*}\limits  f^t(\w') \leq \maxdualitygap \}$ }
\STATE{ {\bfseries Return:} $\w^{t+1}$ }
\end{algorithmic}
\end{algorithm}

\vspace{-2mm}
\subsection{Independent Online Learning --- \IOCP}
As a baseline, we consider the approach of independent learning, \emph{i.e.} 
there is no coordination among the learners.
To keep track of how often task $\z$ has been observed until time $t$, we introduce $\counter^t_\z = \sum^t_{s=1} \mathds{1}_{\{\z^s = \z\}}$. Each learner $\OCP_\z$ for task $\z \in \problems$ maintains an individual learning rate proportional to $\sfrac{1}{\sqrt{\counter^t_\z}}$ and performs one gradient update step using Equation~\eqref{eq.OCP-update} whenever $\z^t = \z$. The regret of \IOCP is bounded as follows. 
\vspace{-0.8mm}
\begin{theorem} \label{theorem:IOCP}
For individual learning rates $\eta^t_\z = \frac{\eta}{\sqrt{\counter^t_\z}}$ where $\eta = \frac{\norm{\solutionspace_{max}}}{\norm{\maxgradient_{max}}}$, the regret of \IOCP is bounded as
\begin{align*}
\regret_{\IOCP}(T) \leq \frac{3}{2} \sqrt{T \problems} \norm{\solutionspace_{max}} \norm{\maxgradient_{max}}. 
\end{align*}
\end{theorem}

\begin{algorithm}[!t]
	\caption{Learner $\OCP_\z$ for task $\z$}\label{algo.OCP}
\begin{algorithmic}[1]
	\STATE{ {\bfseries Input:}      
	    \begin{itemize}
    			\item Coordination steps: $(\xi^t)_{t \in [T]} \textnormal{ where } \xi^t \in \{0,1\}$ 
    			\item Learning rate constant: $\eta > 0$ 
	    \end{itemize}
    }
	\STATE{	{\bfseries Initialize:}  {$\w^1_\z \in \solutionspace_\z$, $\counter^0_\z = 0$} }
	\FOR{$t = 1, 2, \ldots, T$}
		\IF{$\z^t = \z$}
			\STATE{ Suffer loss $\loss^t(\w^t_{\z})$; Calculate (sub-)gradient $\gradient^t_{\z}$ }
			\STATE{ Update $\counter^t_{\z} = \counter^{t-1}_{\z} + 1$; $\wtilde^{t+1}_{\z} = \w^t_{\z} - \frac{\eta}{\sqrt{\counter^t_{\z}}} \gradient^t_{\z}$}
			\IF{$\xi^t = 1$}
				\STATE{ $\wtilde^{t+1}_\z, \counter^t_{\z} \rightarrow$ \texttt{SHARE} \texttt{WITH} \COCP }
				\STATE{ $\w^{t+1}_\z \ \ \ \ \gets$ \texttt{RECEIVE} \texttt{FROM} \COCP	}
			\ELSE
				\STATE{ $\w^{t+1}_\z = \argmin_{\w \in \solutionspace_\z}  \normnew{\w - \wtilde^{t+1}_\z}_2$	}
			\ENDIF
		\ELSE
			\STATE{ $\counter^t_{\z} = \counter^{t-1}_{\z}$; $\w^{t+1}_\z = \w^{t}_{\z}$ }
			\IF{$\xi^t = 1$}
				\STATE{ $\w^{t+1}_\z, \counter^t_{\z} \rightarrow$ \texttt{SHARE} \texttt{WITH} \COCP }
				\STATE{ $\w^{t+1}_\z \ \ \ \ \gets$ \texttt{RECEIVE} \texttt{FROM} \COCP}
			\ENDIF
		\ENDIF
	\ENDFOR
\end{algorithmic}
\end{algorithm}
\subsection{Coordinated Online Learning --- \COCP}
Now, we present our methodology for coordinating these individual online learners. 
Our proposed algorithm -- \COCP\ -- playing the role of a central coordinator is given in Algorithm~\ref{algo.COCP}, and the algorithm of the individual learners $\OCP_\z$ according to the above-mentioned specification is given in Algorithm~\ref{algo.OCP}. 
The \COCP algorithm makes use of a function \aprxproj (Function~\ref{algo.aprxproj}) for computing approximate projections.
\subsubsection{High-level Overview}
The execution of the Algorithm~\ref{algo.COCP} is defined by two parameters: (i) the sequence of coordination steps $(\xi^t)_{t \in [T]}$ where $\xi^t = 0$ means that no coordination happens at time $t$ and (ii) the sequence $(\dualitygap^t)_{t \in [T]}$ whereby $\dualitygap^t$ denotes the desired accuracy of projection at time $t$. 
In Algorithm~\ref{algo.COCP}, the sequences $(\dualitygap^t)_{t \in [T]}$ and $(\xi^t)_{t \in [T]}$ are given as input, however the algorithm \COCP could also set the  values of $\xi^t$ or $\dualitygap^t$ dynamically. Our methodology operates in a synchronized way, in a sense that \COCP can coordinate with the learners at any time $t$: for clarity of presentation, we provide $(\xi^t)_{t \in [T]}$ as input to the learners $\OCP_\z$ as presented in Algorithm~\ref{algo.OCP}. The communication between the \COCP algorithm and learners is represented by $\texttt{RECEIVE}$ and $\texttt{SHARE}$ commands, \emph{cf.} Algorithm~\ref{algo.COCP} and Algorithm~\ref{algo.OCP}. At time $t$ when $\xi^t = 1$, the algorithm \COCP $\texttt{RECEIVE}$s (Line~$4$) the current weight vectors $\w_\z$ and counters $\counter^t_{\z}$ from the learners. And, \COCP $\texttt{SHARE}$s (Line~$8$) the updated weight vectors $\w^{t+1}_\z$ obtained via coordination. In the rest of this section, we will discuss the key ideas used in the development of our algorithm.
\subsubsection{Coordination via Weighted Projection}
Our goal is to design efficient algorithms to coordinate the learners to exploit the tasks relationship modeled by the convex set $\solutionspace^*$. The key question to address is: At time $t$, how can we \emph{aggregate} the current weight vectors $\w^t_\z \ \forall \z \in \problems$ of the individual learners to exploit the structure among the tasks? As shown in the Appendix, a principled way to coordinate in this setting is via performing a  \emph{weighted projection} to $\solutionspace^*$, with \emph{weights} for a learner $\OCP_\z$ being proportional to $\sqrt{\counter^t_\z}$.

We define $Q^t$ as a square diagonal matrix of size $d \problems$ with each $\sqrt{\counter^t_\z}$ represented $d$ times. In the one-dimensional case ($d=1$), we can write $Q^t$ as
\begin{align}
\Q^{t} = 
\begin{bmatrix}
\sqrt{\counter^t_1} & & 0  \\
& \ddots \\
0 & & \sqrt{\counter^t_\problems}
\end{bmatrix}.
\label{eq.qmatrix}
\end{align}
Using $\wtilde$ to jointly represent the current  weight vectors of all the learners at time $t$ (\emph{cf.}\ Line~$5$ in Algorithm~\ref{algo.COCP}), we compute the new joint weight vector $\w^{t+1}$ (\emph{cf.}\ Line~$7$ in Algorithm~\ref{algo.COCP}) by projecting onto $\solutionspace^*$, using the squared Mahalanobis distance, \emph{i.e.}
\begin{align}
\w^{t+1} &= \argmin_{\w \in \solutionspace^*} (\w - \wtilde)' \Q^t (\w - \wtilde). \label{eq.weightedproj}
\end{align}

We refer to this as the weighted projection onto $\solutionspace^*$ and note that since $\solutionspace^*$ is convex, the projection is unique: the weighted projection is a special case of the Bregman projection\footnote{For shared parameters, the weighted projection is equivalent to the weighted average of the parameters.}, \emph{cf.}\ Appendix and \cite{cesa2006prediction,rakhlin2009lecture}.
Intuitively, the weighted projection allows us to learn about tasks that have been observed infrequently, while avoiding to ``unlearn" about the tasks that have been observed more frequently.  Algorithm~\ref{algo.COCP}, when invoked with $\xi^t = 1, \dualitygap^t = 0  \ \forall t \in [T]$, corresponds to a variant of our algorithm that does exact/noise-free coordination at every time step. When invoked with $\xi^t = 0 \ \forall t \in [T]$, our algorithm corresponds to the \IOCP baseline.
\subsubsection{Sporadic \& Approximate Coordination}
For large scale applications (\emph{i.e.} large $\problems$ or large $d$), coordination at every step by performing projections could be computationally very expensive: a projection onto a generic convex set $\solutionspace^*$ would require solving a quadratic program of dimension $d \cdot \problems$. Furthermore, it is desirable to design a communication-efficient coordination protocol when applying our framework in a distributed setting. We extend our approach with two novel algorithmic ideas: The \COCP algorithm can perform sporadic and approximate coordinations, defined by the above-mentioned sequences $(\xi^t)_{t \in [T]}$  and $(\dualitygap^t)_{t \in [T]}$. 
Here, $\dualitygap^t$ denotes the desired accuracy or the amount of noise that is allowed at time $t$, and is given as input to the function \aprxproj (Function~\ref{algo.aprxproj}) for computing approximate projections. As we shall see in our experimental results, these two algorithmic ideas of sporadic and approximate coordination allows us to speed up the algorithm by an order of magnitude while retaining the improvements obtained by coordination. Furthermore, these two features have privacy-preserving properties that can be exploited in the distributed setting, if privacy or data leakage across the learners is a concern \cite{balcan2012distributed}. For instance by using ideas of differential privacy in online learning (\emph{cf.} \cite{jain2012differentially}), the algorithm can define the desired $\dualitygap^t$ at time $t$ and perturb the projected solution by noise level $\dualitygap^t$.
\subsubsection{Remarks}
We conclude the presentation of our methodology with a few remarks. 
We note that the \COCP algorithm bears resemblance to the adaptive gradient based algorithms, such as \AdaGrad \cite{duchi2011adaptive}. In a completely centralized setting, another way to view this online multi-task learning problem is to treat each task as representing $d$ parameters/features of the joint online learning problem with dimension $d \cdot K$. Then, at each time $t$, observing an instance of task $\z^t$ is equivalent to receiving a sparse (sub-)gradient with only up to $d$ non-zero entries corresponding to the features of task $\z^t$. In fact, we can formally show that for $d=1$, when the (sub-)gradients $\gradient^t_{\z^t} \in \{-1, 1\} \ \forall t \in [T]$, the learning behavior of \AdaGrad (with $\solutionspace^*$ as the feasible solution space) and \COCP (with $\xi^t = 1, \dualitygap^t = 0  \ \forall t \in [T]$) are equivalent. However, our methodology is more widely applicable whereby each task is being tackled by a separate online learner, and furthermore allows us to perform sporadic/approximate coordination by controlling  $\xi^t$ and $\dualitygap^t$.
\section{Theoretical Guarantees}\label{sec.analysis}
In this section, we analyze the regret bounds of the \COCP algorithm; all proofs are provided in Appendix.

\subsection{Generic Bounds}
We begin with a general result in Theorem~\ref{theorem:COCP} and then we will refine these bounds for specific settings. 
\begin{theorem} \label{theorem:COCP}
The regret of the \COCP algorithm is bounded by $\regret_{\COCP}(T) \leq$
\begin{align}
&\quad \frac{1}{2 \eta} \norm{\solutionspace_{max}}^2 \sqrt{T \problems} +  2 \eta \norm{\maxgradient_{max}}^2 \sqrt{T\problems} \label{theorem:COCP.term1}  \tag{R1} \\
&\quad+  \sum^T_{t=1} \bad \norm{\solutionspace_{max}}  \norm{\maxgradient_{max}} \label{theorem:COCP.term2} \tag{R2} \\
&\quad+ \frac{1}{\eta} \sum^T_{t=1}  \inside \left( \maxdualitygap + \sqrt{2 \maxdualitygap}(t\problems)^{1/4} \norm{\solutionspace_{max}} \right) \label{theorem:COCP.term3} \tag{R3} \\
&\quad+ \frac{1}{2 \eta} \norm{\solutionspace_{max}}^2  - 2 \eta \norm{\maxgradient_{max}}^2 \problems. \label{theorem:COCP.term4} \tag{R4}
\end{align}
\end{theorem}
Intuitively, the regret in Theorem~\ref{theorem:COCP} has four components. Sporadic coordination leads to \ref{theorem:COCP.term2} and allowed noise in the projection leads to \ref{theorem:COCP.term3}.  \ref{theorem:COCP.term1} comes from the standard regret analysis and \ref{theorem:COCP.term4} is a constant.

\subsection{Sporadic/Approx. Coordination Bounds}
In our experiments, \emph{cf.} Section~\ref{sec.experiments}, we consider the setting where \COCP  projects onto $\solutionspace^*$ approximately and with a low probability, attempting to combine the benefits of the coordination with a low average computational complexity. The regret bounds for this practically useful setting are stated in Corollary~\ref{corollary:COCP.rare}.
\begin{corollary} \label{corollary:COCP.rare}
Set $\eta = \frac{1}{2} \frac{\norm{\solutionspace_{max}}}{\norm{\maxgradient_{max}}}$. $\forall t \in [T]$, define:
\begin{align*}
\xi^t \sim Bernoulli(\alpha) \textnormal{ with } \alpha = \frac{\calpha}{\sqrt{T}}, \\
\maxdualitygap = \cbeta (1 - \beta)^2 \frac{\sqrt{\problems}}{\sqrt{t}} \norm{\solutionspace_{max}}^2
\end{align*}
where constants $\calpha \in [0, \sqrt{T}]$, $\cbeta \geq 0$, and $\beta \in [0,1]$. 
The expected regret of \COCP (where the expectation is w.r.t. $(\xi^t)_{t\in[T]}$) is bounded by $\expectedvalue{\regret_{\COCP}(T)} \leq$
\begin{align*}
&\quad 2 \sqrt{T \problems} \norm{\solutionspace_{max}}  \norm{\maxgradient_{max}} \cdot \bigg(1 + \frac{\calpha}{2\sqrt{\problems}}  \Big(1 - \frac{\calpha}{\sqrt{T}} \Big)& \\
& \qquad \qquad \qquad \qquad \qquad \quad + \calpha (\cbeta + \sqrt{2 \cbeta}) (1 - \beta) \bigg).&
\end{align*}
\end{corollary}

\subsection{Bounds for Specific Settings}
Algorithm~\ref{algo.COCP} when invoked with $\xi^t = 1, \dualitygap^t = 0  \ \forall t \in [T]$ does exact/noise-free coordination at every time step. 
For this special case, we refine the bounds of Theorem~\ref{theorem:COCP} in the following corollary.

\begin{figure*}[!t]
\centering
  \begin{subfigure}[b]{0.31\textwidth}
    \centering
    \includegraphics[trim = 2.5mm 3.2mm 4mm 10.5mm, clip=true, width=1\textwidth]{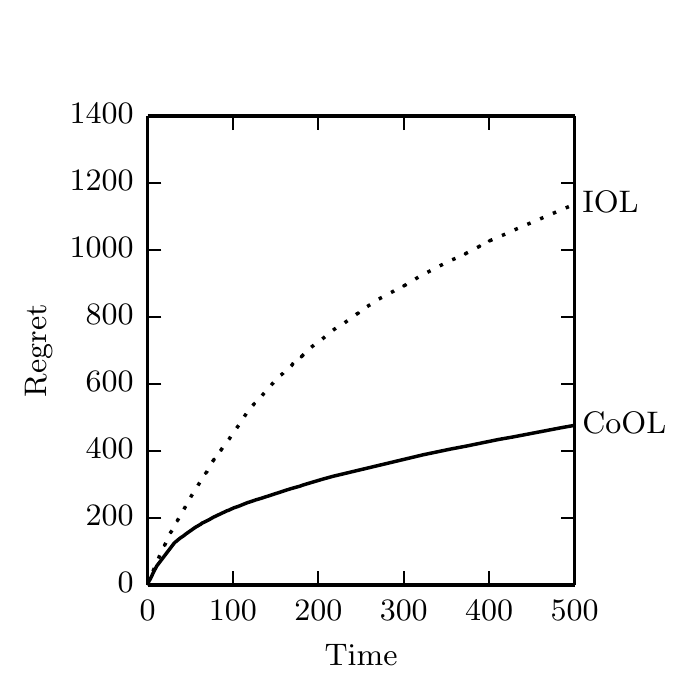}
    \caption{Random order of tasks}
    \label{fig.random}
  \end{subfigure}
  \begin{subfigure}[b]{0.31\textwidth}
    \centering
    \includegraphics[trim = 2.5mm 3.2mm 4mm 10.5mm, clip=true, width=1\textwidth]{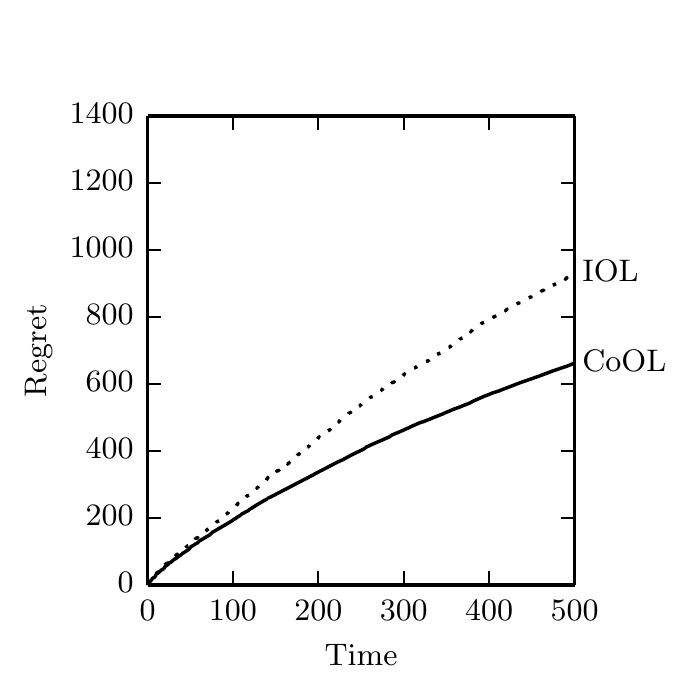}
    \caption{Batches of tasks}
    \label{fig.batch}
  \end{subfigure}
  \begin{subfigure}[b]{0.31\textwidth}
    \centering
    \includegraphics[trim = 2.5mm 3.2mm 4mm 10.5mm, clip=true, width=1\textwidth]{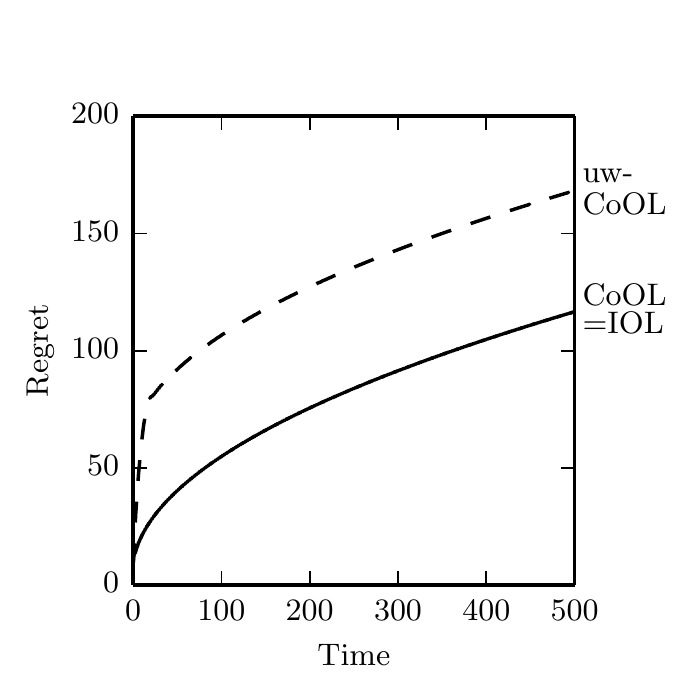}
    \caption{Single task}
    \label{fig.single}
  \end{subfigure}
  \\
  \begin{subfigure}[b]{0.31\textwidth}
    \centering
    \includegraphics[trim = 2.5mm 3.2mm 4mm 10.5mm, clip=true, width=1\textwidth]{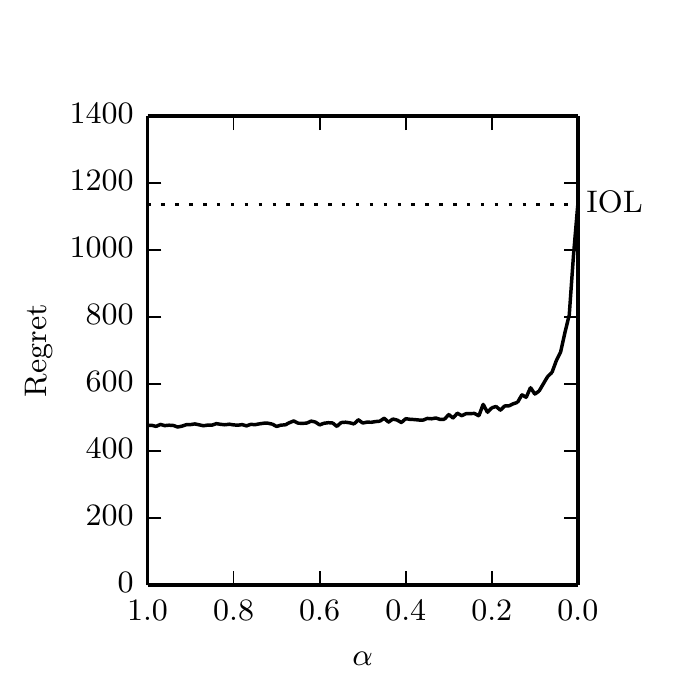}
    \caption{Varying $\alpha$}
    \label{fig.alpha}
  \end{subfigure}
  %
  \begin{subfigure}[b]{0.31\textwidth}
    \centering
    \includegraphics[trim = 2.5mm 3.2mm 4mm 10.5mm, clip=true, width=1\textwidth]{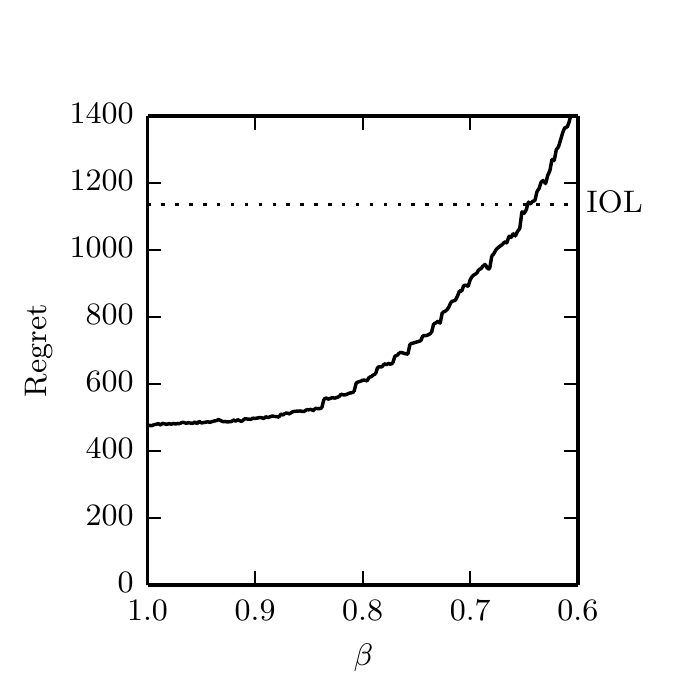}
    \caption{Varying $\beta$}
    \label{fig.beta}
  \end{subfigure}
  %
  \begin{subfigure}[b]{0.31\textwidth}
    \centering
    \includegraphics[trim = 2.5mm 3.2mm 4mm 10.5mm, clip=true, width=1\textwidth]{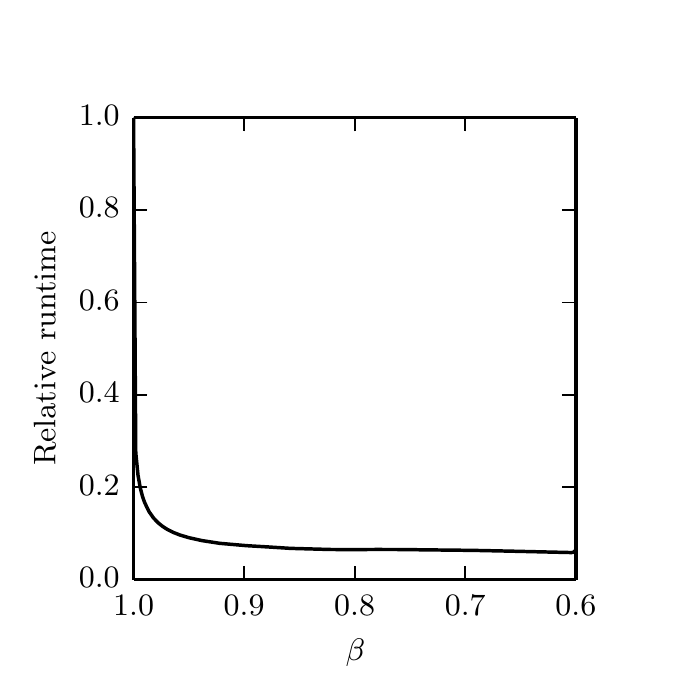}
    \caption{Runtime when varying $\beta$}
    \label{fig.time}
  \end{subfigure}      
\caption{Simulation results for learning hemimetrics. (a,b,c) compare the performance of \COCP against \IOCP for different orders of task instances. (d,e,f)  show the tradeoffs in performing sporadic/approximate coordination.}
\label{fig.simulations}
\end{figure*}

\begin{corollary} \label{corollary:COCP.special}
Set $\eta = \frac{1}{2} \frac{\norm{\solutionspace_{max}}}{\norm{\maxgradient_{max}}}$ and  $\xi^t = 1, \dualitygap^t = 0  \ \forall t \in [T]$. Then, the regret of \COCP is bounded by
\begin{align*}
\regret_{\COCP}(T) \leq 2 \sqrt{T \problems} \norm{\solutionspace_{max}}  \norm{\maxgradient_{max}}.
\end{align*}
A more careful analysis for this case and by using the learning rate constant $\eta = \frac{\norm{\solutionspace_{max}}}{\norm{\maxgradient_{max}}}$ gives a tighter regret bound with a constant factor $\frac{3}{2}$ instead of $2$. This matches the regret bound of \IOCP stated in Theorem~\ref{theorem:IOCP}.
\end{corollary}
This corollary shows that our approach to coordinate via weighted projection using weights as in Equation~\eqref{eq.qmatrix} preserves the worst-case guarantees of the \IOCP baseline algorithm. We further illustrate in the experiments (\emph{cf.} Figure~\ref{fig.single}) that using the wrong weights (\emph{e.g.} un-weighted projection) can hinder the convergence of the learners. 

This corollary also reveals the worst-case nature of the regret bounds proven in Theorem~\ref{theorem:COCP}, \emph{i.e.} the proven bounds for \COCP are agnostic to the specific structure $\solutionspace^*$ and the order of task instances. However, for some specific setting we can get better bounds for the \COCP algorithm.
For instance, in the following theorem we consider a fixed $\epsilon$-insensitive loss function and a $B$-batch order of tasks, where a task instance is repeated $B$ times before choosing a new task instance.

%
%
%
\begin{theorem} \label{theorem:COCP-stochastic}
Consider $d=1$ with shared parameter structure, and $\epsilon$-insensitive loss function given by $\loss^t(\w^t_{\z^t}) = 0  \textnormal{ if } \lvert \w^t_{\z^t} - c^* \rvert \leq \epsilon$, else  $\loss^t(\w^t_{\z^t}) = \lvert \w^t_{\z^t} - c^* \rvert - \epsilon$, where $\epsilon > 0$ and $c^* \in \mathbb{R}$ is a constant. For $\eta = \frac{\norm{\solutionspace_{max}}}{\norm{\maxgradient_{max}}}$, in the $B$-batch setting with sufficiently large batch size $B \geq \big\lceil(\frac{\norm{\solutionspace_{max}}}{\epsilon} + \frac{1}{2})^2\big\rceil$, the regret of the \COCP algorithm is bounded by
\begin{align*}
\regret_{\COCP}(T) \leq \frac{3}{2} \sqrt{B} \norm{\solutionspace_{max}}  \norm{\maxgradient_{max}},
\end{align*}
whereas the regret bound of the \IOCP algorithm is worse by up to a factor $\problems$.
\end{theorem}

\section{Experimental Evaluation}\label{sec.experiments}

{\bfseries Learning hemimetrics.} Our simulation experiments are based on learning hemimetrics, \emph{cf.} motivating applications in Section~\ref{sec.introduction.app}. 
We consider $d=1$ and model the underlying structure $\solutionspace^*$ as a set of $r$-bounded hemimetrics, \emph{cf.} Section~\ref{subsec.relatedness}.
Similar to \cite{singla2016actively}, 
we generated an underlying \emph{ground-truth} hemimetric $\w^*$ from a clustered setting where the $n$ items belong to two equal-sized clusters. We define the distance $D_{i,j} = r_{in}$ if $i$ and $j$ are from the same cluster, and $D_{i,j} = r_{out}$ otherwise. In the experiments, we set $n=10$ (resulting in $\problems = 90$), $r_{in} = 1$, and $r_{out} = 9$.

{\bfseries Loss function and gradients.} 
For a given instance of task $\z^t$ at time $t$, the \emph{offer} is represented by the prediction $\w^t_{\z^t}$ from learner $\OCP_{\z^t}$. We consider a simple loss function given by $l^t(\w^t_{\z^t}) = \lvert \w^t_{\z^t} - \w^*_{\z^t} \rvert$. When the offer $\w^t_{\z^t} \geq \w^*_{\z^t}$, the user ``accepts" and the gradient $\gradient^t_{\z^t} = +1$; otherwise $\gradient^t_{\z^t} = -1$. This interaction with the users is motivated by the \emph{posted-price model} in marketplaces  \cite{abernethy2015low,singla2013truthful},  where users are offered a take-it-or-leave-it price, which they can accept or reject.

{\bfseries Projection algorithm.} For computing approximation projections in \aprxproj (Function~\ref{algo.aprxproj}), we adapted the \emph{triangle fixing} algorithm proposed for the \emph{Metric Nearness Problem} in \cite{brickell2008metric}. While the original algorithm was designed for performing unweighted/exact projections for metrics, we adapted it to our setiting of weighted projections to hemimetrics with additional ability to perform approximate projection controlled by the duality gap.  
 

\subsection{Results: Order of Task Instances}\label{sec.experiments.results}
In this set of experiments, we consider the \COCP algorithm with exact/noise-free coordination (\emph{i.e.} $\alpha=1, \beta=1$ in Corollary~\ref{corollary:COCP.rare}) at every time step. We compare the effect of the order and the number of different tasks instances received --- the results are shown in Figure~\ref{fig.simulations}, averaged over 10 runs.

{\bfseries Random order of tasks.} Task instances $\z^t$ are chosen uniformly at random at every time step. The \COCP algorithm suffers a significantly lower regret than the \IOCP algorithm, benefiting from the weighted projection onto $\solutionspace^*$. At $T=500$, the regret of \COCP is less than half of that of the \IOCP, \emph{cf.} Figure~\ref{fig.random}.

{\bfseries Batches of tasks.} In the batch setting, a task instance is  chosen uniformly at random, then it is repeated five times before choosing a new task instance. The \IOCP algorithm suffers a lower regret compared to the above-mentioned random order because  of the higher probability that certain tasks are shown a large number of times. Furthermore, the benefit of the projection onto $\solutionspace^*$ for the  \COCP algorithm is reduced, \emph{cf.} Figure~\ref{fig.batch}, showing that the benefit of coordination depends on the specific order of the task instances for a given structure.

{\bfseries Single-task setting.} A single task $\z$ is repeated in every round. In this case, the \IOCP algorithm and the \COCP algorithm have same regret as illustrated, \emph{cf.} Figure~\ref{fig.single}. In order to get better understanding  of using weights $Q^t$ for the weighted projection, we also show a variant $\text{uw-}\COCP$ using $Q^t$ as identity matrix. Un-weighted projection or using the wrong weights can hinder the convergence of the learners, as shown in Figure~\ref{fig.single} for this extreme case of a single-task setting.


\subsection{Results: Rate/Accuracy of Coordination}
Next, we compare the trade-offs of computation vs. benefits from coordination via sporadic/approximate coordination, by varying $\alpha, \beta$ in Corollary~\ref{corollary:COCP.rare}.

{\bfseries Varying the rate of coordination (\boldmath$\alpha$).} The regret of the \COCP algorithm monotonically increases as $\alpha$ decreases, and is equivalent to the regret of the \IOCP algorithm when $\alpha=0$, \emph{cf.} Figure~\ref{fig.alpha}. In the range of $\alpha$ values between $1$ and $0.1$, the regret of the \COCP algorithm is relatively constant and increases strongly only as $\alpha$ approaches $0$. With $\alpha$ as low as $0.1$, the regret of the \COCP algorithm in this setting is still almost half of that of the \IOCP algorithm. 

{\bfseries Varying the accuracy of coordination (\boldmath$\beta$).} The regret of the \COCP algorithm monotonically increases as $\beta$ decreases, and exceeds that of the \IOCP algorithm for values smaller than $0.65$ because of high noise in the projections, \emph{cf.} Figure~\ref{fig.beta}. In the range of $\beta$ values between $1$ and $0.85$, the regret of the \COCP algorithm is  relatively constant and less than half of that of the \IOCP algorithm.

{\bfseries Runtime vs. approximate projections.} As expected, the runtime of the projection monotonically decreases as $\beta$ decreases, \emph{cf.} Figure~\ref{fig.time}. For values of $\beta$ smaller than $0.95$, the runtime of the projection is less than $10\%$ of that of the exact projection. Thus, with $\beta$ values in the range of $0.85$ to $0.95$, the \COCP algorithm achieves the best of both the worlds: the regret is significantly smaller than that of \IOCP, with an order of magnitude speedup in the runtime.
\begin{figure*}[!t]
\centering
  %
  \begin{subfigure}[b]{0.30\textwidth}
    \centering
    \includegraphics[trim = 0mm 0mm 0mm 25mm, clip=true, width=1\textwidth]{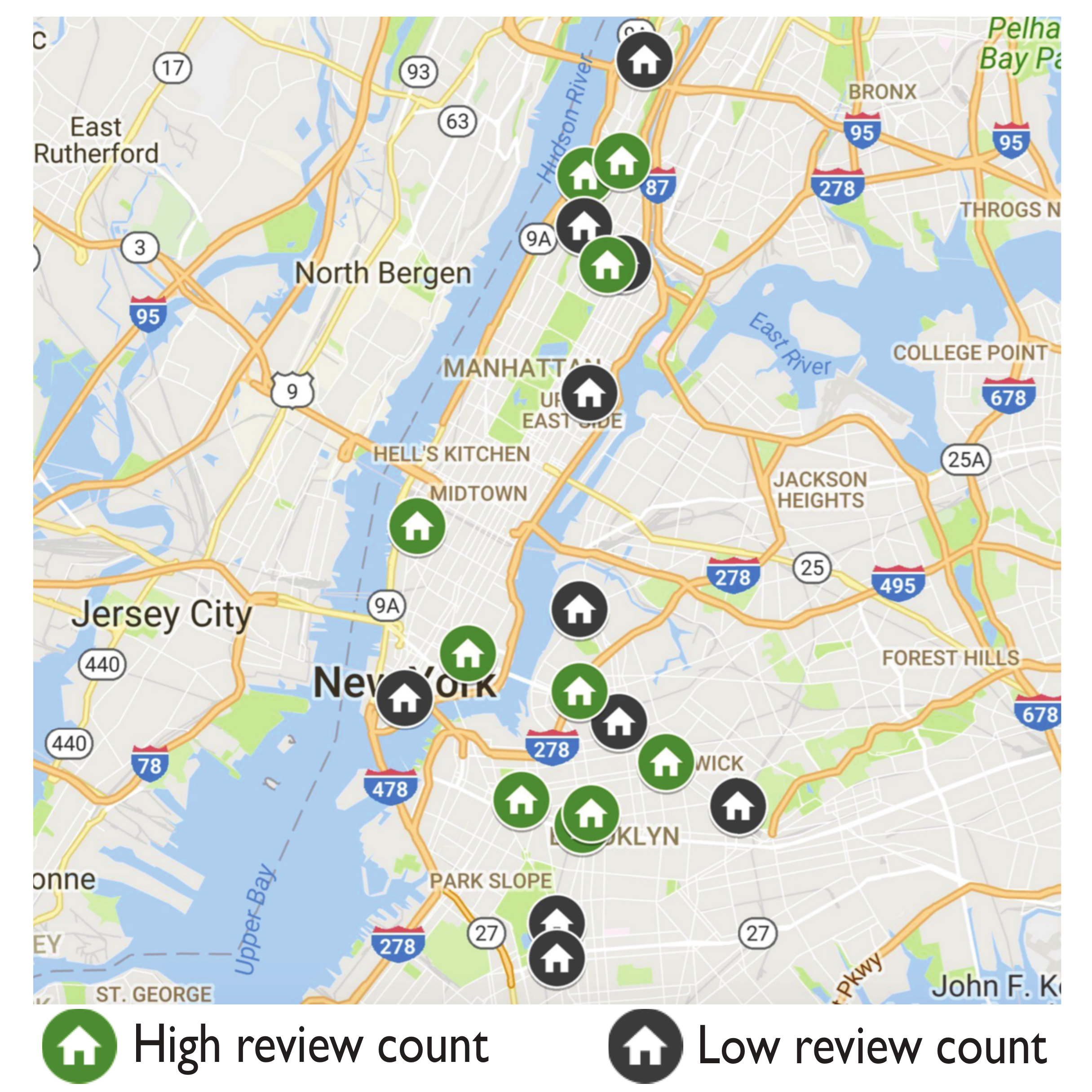}
    \caption{Dataset of 20 items}
    \label{fig.map}
  \end{subfigure}
  %
  \begin{subfigure}[b]{0.29\textwidth}
    \centering
    \includegraphics[trim = 2.6mm 3.2mm 1.5mm 2mm, clip=true, width=1\textwidth]{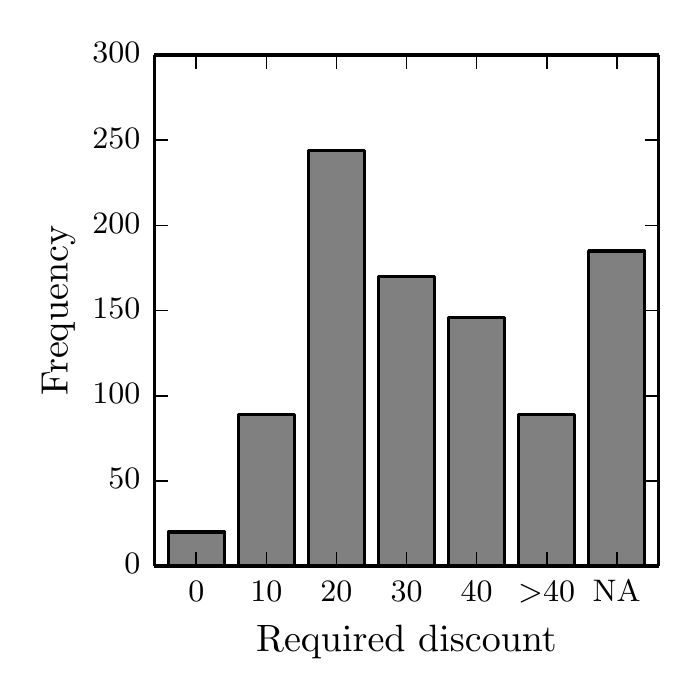}
    \caption{Distribution of elicited costs}
    \label{fig.discounts}
  \end{subfigure}
  %
  \begin{subfigure}[b]{0.31\textwidth}
    \centering
    \includegraphics[trim = 2.6mm 3.2mm 1.5mm 2mm, clip=true, width=1\textwidth]{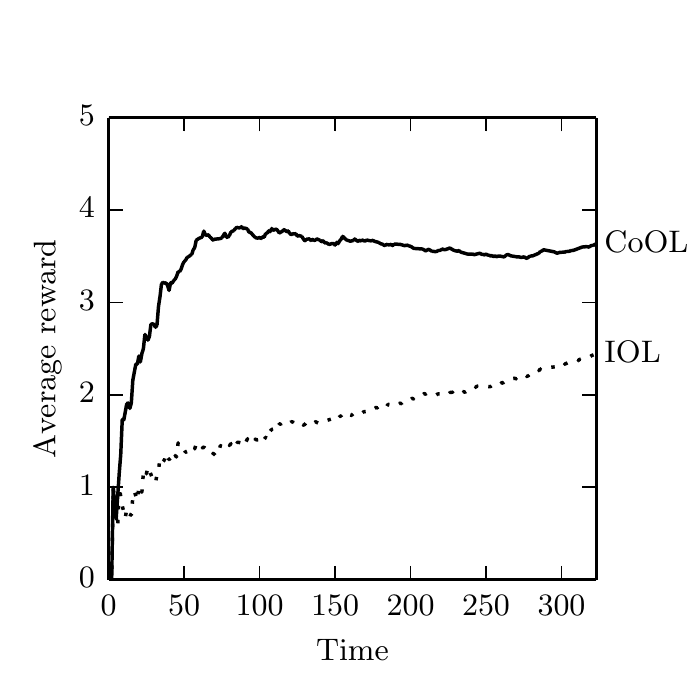}
    \caption{Average reward / convergence}
    \label{fig.user_highlow_avg}
  \end{subfigure}
  %
\caption{Results based on \names{Airbnb} dataset and survey study for \names{Airbnb} marketplace}
\label{fig.realworld}
\end{figure*}



\section{Case Study on \names{Airbnb} Marketplace}\label{sec.userstudy}
We now study the problem of learning users' preferences on \names{Airbnb} with the goal of incentivizing users to explore  under-reviewed apartments \cite{kamenica2009bayesian,singla2016actively}.

{\bfseries \names{Airbnb} dataset.}
Using data of \names{Airbnb} apartments from \names{insideairbnb.com} \cite{insideairbnb}, we created a dataset of 20 apartments as follows. We  chose apartments from $4$ types in the New York City: (i) based on location (Manhattan or Brooklyn) and (ii) the number of reviews (high, $\geq 20$ or low, $\leq 2$). From each type we chose 5 apartments, resulting in a total sample of $n=20$ apartments, displayed in Figure~\ref{fig.map}. 

{\bfseries Survey study on MTurk platform.} In order to get real-world distributions of the users' private costs, we collected data from Amazon's Mechanical Turk \cite{mturk} as follows. Each participant was presented two randomly chosen apartments and asked to select her preferred choice (\emph{cf.} Appendix for a snapshot). Participants were then asked to specify their private cost for switching their choice to the other apartment. The resulting dataset consists of tuples $((i,j),c)$, where $i$ is the preferred choice, $j$ is the suggested choice, and $c$ is the private cost of the user. In total, we got $943$ responses/tuples. The distribution of elicited costs is shown in Figure~\ref{fig.discounts}, where \textnormal{NA} corresponds to about $20\%$ participants who were unwilling to accept any offer. In $323$ responses $i$ was a high-reviewed apartment, $j$ an under-reviewed apartment, and participants did not select \textnormal{NA}. We use these responses in our experiments as explained below. 

{\bfseries Utility/rewards.}
A time step $t$ corresponds to a tuple $((i^t, j^t), c^t)$ with task instance $\z^t = (i^t, j^t)$, and we have $T=323$.
Let $p^t$ denote the offered price by learner $\OCP^t_{\z^t}$ based on the current weight vector $\w^t_{\z^t}$. We model the utility and reward of the marketplace as follows. The reward at time $t$ is $(u - p^t)$ if the offer $p^t$ is accepted (\emph{i.e.} $p^t \geq c^t$), and otherwise zero. Here, $u$ is the utility of the marketplace for getting a review for an under-reviewed apartment, and is set to $u=40$ in our experiments based on referral discounts given by the marketplace in past. We can model the above-mentioned rewards by the following (discontinuous) loss function: $l^t(p^t) = \mathds{1}_{\{p^t \geq c^t\}} \cdot (p^t - c^t) + \mathds{1}_{\{p^t < c^t\}} \cdot (u - c^t)  \textnormal{ for } u \geq c^t$, and $l^t(p^t) = 0 \textnormal{ for } u < c^t$.

{\bfseries Loss function and gradients.} For running the experiments, we consider a simple convex loss function given by $l^t(p^t) = \mathds{1}_{\{p^t \geq c^t\}} \cdot (p^t - c^t) + \mathds{1}_{\{p^t < c^t\}} \cdot \frac{u}{\Delta} \cdot (c^t - p^t)$ where $\frac{u}{\Delta}$ denotes the magnitude of the gradient when a user rejects the offer, where the value of parameter $\Delta$ is set to $20$ in the experiments. Using this loss function also allows us to compute the gradients from binary feedback of acceptance/rejection of the offers.

\subsection{Results}
We have a total of $\problems = n^2 - n$ learning tasks with $n=20$ items. Similar to Section~\ref{sec.experiments}, we consider $d=1$ and use a hemimetric structure to model the relationship of the tasks. The results of this experiments are shown in Figure \ref{fig.user_highlow_avg} showing the average reward per time step and a faster convergence of the \COCP algorithm compared to that of the \IOCP algorithm. 
\section{Related Work}\label{sec.related}
{\bfseries Online/distributed multi-task learning.}
Multi-task learning has been increasingly studied in online and distributed settings recently. 
Inspired by wearable computing, a recent work by \cite{jin2015collaborating} studied online multi-task learning in a distributed setting. They considered a setup where tasks arrive asynchronously and the relatedness among the tasks is maintained via a correlation matrix.  However, there is no theoretical analysis on the regret bounds for the proposed algorithms. \cite{wang2016distributed} recently studied the multi-task learning for distributed LASSO with shared support. Their work is different from ours --- we consider general convex constraints to model task relationships and consider the adversarial online regret minimization framework.

{\bfseries Modeling task relationships.}
Similar in spirit to ours, some previous work has focused on general frameworks to model task relationships. \cite{dekel2007online} models this via a global loss function that combines the loss values of the individual tasks incurred at a given time. This global loss function is restricted to a family of absolute norms. \cite{lugosi2009online} models the task relationships by enforcing a set of hard constraints on the joint action space of the tasks and restrict these constraints to satisfy a Markovian property for computational efficiency.
One key difference compared to \cite{dekel2007online,lugosi2009online} is that in our work, tasks are not required to be executed simultaneously at a given time, making it applicable to distributed learning of the tasks. \cite{abernethy2007multitask} studies online multi-task learning in the framework of prediction with expert advice by restricting the number of ``best" experts. Another line of work, complementary to ours, considers learning the task relationships jointly with learning the tasks themselves  \cite{kang2011learning,saha2011online,ciliberto2015convex}. 



%

{\bfseries Distributed optimization.}
Our results have some similarity with consensus problems in the distributed (stochastic) optimization literature \cite{boyd2011distributed,dekel2012optimal,shamir2014distributed}. \cite{nedic2009distributed,yan2013distributed} study the problem of distributed autonomous online learning where each learner has its own sequence of loss functions. These learners can communicate on a network with their neighbors to share their parameters.
Distributed consensus problems can be viewed as distributed multi-task problems with a constraint structure of some parameters being shared among the tasks. Our approach is applicable to these problems as long as centralized coordination is possible.
\section{Conclusions}\label{sec.conclusions}
We studied online multi-task learning by modeling the relationship of tasks via a set of convex constraints. To exploit this relationship, we developed a novel algorithm, \COCP, to coordinate the task-specific online learners. The key idea of our algorithm for coordination is to perform weighted projection of the current solution vectors of the learners onto a convex set. Furthermore, \COCP can perform sporadic and approximate coordinations, thereby making it suitable for real-world applications where computation complexity is a bottleneck or low-communication/privacy is important. Our theoretical analysis yields insights into how these trade-off factors influence the regret bounds. Our experimental results on \names{Airbnb} demonstrate the practical applicability of our approach.
\bibliography{refs}

\begin{thebibliography}{32}
\providecommand{\natexlab}[1]{#1}
\providecommand{\url}[1]{\texttt{#1}}
\expandafter\ifx\csname urlstyle\endcsname\relax
  \providecommand{\doi}[1]{doi: #1}\else
  \providecommand{\doi}{doi: \begingroup \urlstyle{rm}\Url}\fi

\bibitem[ins()]{insideairbnb}
{I}nside {A}irbnb.
\newblock \url{http://insideairbnb.com/}.

\bibitem[mtu()]{mturk}
{M}echanical {T}urk.
\newblock \url{https://www.mturk.com/}.

\bibitem[Abernethy et~al.(2015)Abernethy, Chen, Ho, and
  Waggoner]{abernethy2015low}
Abernethy, J., Chen, Y., Ho, C., and Waggoner, B.
\newblock Low-cost learning via active data procurement.
\newblock In \emph{EC}, pp.\  619--636, 2015.

\bibitem[Abernethy et~al.(2007)Abernethy, Bartlett, and
  Rakhlin]{abernethy2007multitask}
Abernethy, Jacob, Bartlett, Peter, and Rakhlin, Alexander.
\newblock Multitask learning with expert advice.
\newblock In \emph{COLT}, 2007.

\bibitem[Balcan et~al.(2012)Balcan, Blum, Fine, and
  Mansour]{balcan2012distributed}
Balcan, Maria-Florina, Blum, Avrim, Fine, Shai, and Mansour, Yishay.
\newblock Distributed learning, communication complexity and privacy.
\newblock In \emph{COLT}, 2012.

\bibitem[Beckenbach \& Bellman(2012)Beckenbach and
  Bellman]{beckenbach2012inequalities}
Beckenbach, Edwin~F and Bellman, Richard.
\newblock \emph{Inequalities}, volume~30.
\newblock Springer Science \& Business Media, 2012.

\bibitem[Boyd et~al.(2011)Boyd, Parikh, Chu, Peleato, and
  Eckstein]{boyd2011distributed}
Boyd, Stephen, Parikh, Neal, Chu, Eric, Peleato, Borja, and Eckstein, Jonathan.
\newblock Distributed optimization and statistical learning via the alternating
  direction method of multipliers.
\newblock \emph{Foundations and Trends{\textregistered} in Machine Learning},
  2011.

\bibitem[Brickell et~al.(2008)Brickell, Dhillon, Sra, and
  Tropp]{brickell2008metric}
Brickell, Justin, Dhillon, Inderjit~S, Sra, Suvrit, and Tropp, Joel~A.
\newblock The metric nearness problem.
\newblock \emph{SIAM Journal on Matrix Analysis and Applications}, 30\penalty0
  (1):\penalty0 375--396, 2008.

\bibitem[Caruana(1998)]{caruana1998multitask}
Caruana, Rich.
\newblock Multitask learning.
\newblock In \emph{Learning to learn}, pp.\  95--133. Springer, 1998.

\bibitem[Cesa-Bianchi \& Lugosi(2006)Cesa-Bianchi and
  Lugosi]{cesa2006prediction}
Cesa-Bianchi, N. and Lugosi, G.
\newblock \emph{Prediction, learning, and games}.
\newblock Cambridge university press, 2006.

\bibitem[Chapelle et~al.(2010)Chapelle, Shivaswamy, Vadrevu, Weinberger, Zhang,
  and Tseng]{chapelle2010multi}
Chapelle, Olivier, Shivaswamy, Pannagadatta, Vadrevu, Srinivas, Weinberger,
  Kilian, Zhang, Ya, and Tseng, Belle.
\newblock Multi-task learning for boosting with application to web search
  ranking.
\newblock In \emph{KDD}, 2010.

\bibitem[Ciliberto et~al.(2015)Ciliberto, Mroueh, Poggio, and
  Rosasco]{ciliberto2015convex}
Ciliberto, Carlo, Mroueh, Youssef, Poggio, Tomaso, and Rosasco, Lorenzo.
\newblock Convex learning of multiple tasks and their structure.
\newblock In \emph{ICML}, 2015.

\bibitem[Dekel et~al.(2007)Dekel, Long, and Singer]{dekel2007online}
Dekel, Ofer, Long, Philip~M, and Singer, Yoram.
\newblock Online learning of multiple tasks with a shared loss.
\newblock \emph{Journal of Machine Learning Research}, 8:\penalty0 2233--2264,
  2007.

\bibitem[Dekel et~al.(2012)Dekel, Gilad-Bachrach, Shamir, and
  Xiao]{dekel2012optimal}
Dekel, Ofer, Gilad-Bachrach, Ran, Shamir, Ohad, and Xiao, Lin.
\newblock Optimal distributed online prediction using mini-batches.
\newblock \emph{Journal of Machine Learning Research}, 13:\penalty0 165--202,
  2012.

\bibitem[Duchi et~al.(2011)Duchi, Hazan, and Singer]{duchi2011adaptive}
Duchi, John, Hazan, Elad, and Singer, Yoram.
\newblock Adaptive subgradient methods for online learning and stochastic
  optimization.
\newblock \emph{Journal of Machine Learning Research}, 12:\penalty0 2121--2159,
  2011.

\bibitem[Jain et~al.(2012)Jain, Kothari, and Thakurta]{jain2012differentially}
Jain, Prateek, Kothari, Pravesh, and Thakurta, Abhradeep.
\newblock Differentially private online learning.
\newblock In \emph{COLT}, 2012.

\bibitem[Jin et~al.(2015)Jin, Luo, Zhuang, He, and He]{jin2015collaborating}
Jin, Xin, Luo, Ping, Zhuang, Fuzhen, He, Jia, and He, Qing.
\newblock Collaborating between local and global learning for distributed
  online multiple tasks.
\newblock In \emph{CIKM}, 2015.

\bibitem[Kamenica \& Gentzkow(2009)Kamenica and Gentzkow]{kamenica2009bayesian}
Kamenica, Emir and Gentzkow, Matthew.
\newblock Bayesian persuasion.
\newblock Technical report, National Bureau of Economic Research, 2009.

\bibitem[Kang et~al.(2011)Kang, Grauman, and Sha]{kang2011learning}
Kang, Zhuoliang, Grauman, Kristen, and Sha, Fei.
\newblock Learning with whom to share in multi-task feature learning.
\newblock In \emph{ICML}, 2011.

\bibitem[Lugosi et~al.(2009)Lugosi, Papaspiliopoulos, and
  Stoltz]{lugosi2009online}
Lugosi, G{\'{a}}bor, Papaspiliopoulos, Omiros, and Stoltz, Gilles.
\newblock Online multi-task learning with hard constraints.
\newblock In \emph{COLT}, 2009.

\bibitem[Nedic \& Ozdaglar(2009)Nedic and Ozdaglar]{nedic2009distributed}
Nedic, Angelia and Ozdaglar, Asuman.
\newblock Distributed subgradient methods for multi-agent optimization.
\newblock \emph{IEEE Transactions on Automatic Control}, 54\penalty0
  (1):\penalty0 48--61, 2009.

\bibitem[Rakhlin \& Tewari(2009)Rakhlin and Tewari]{rakhlin2009lecture}
Rakhlin, Alexander and Tewari, A.
\newblock Lecture notes on online learning.
\newblock \emph{Draft, April}, 2009.

\bibitem[Saha et~al.(2011)Saha, Rai, III, and
  Venkatasubramanian]{saha2011online}
Saha, Avishek, Rai, Piyush, III, Hal~Daum{\'{e}}, and Venkatasubramanian,
  Suresh.
\newblock Online learning of multiple tasks and their relationships.
\newblock In \emph{AISTATS}, 2011.

\bibitem[Shalev-Shwartz(2011)]{shalev2011online}
Shalev-Shwartz, Shai.
\newblock Online learning and online convex optimization.
\newblock \emph{Foundations and Trends in Machine Learning}, 4\penalty0
  (2):\penalty0 107--194, 2011.

\bibitem[Shamir \& Srebro(2014)Shamir and Srebro]{shamir2014distributed}
Shamir, Ohad and Srebro, Nathan.
\newblock Distributed stochastic optimization and learning.
\newblock In \emph{Allerton}, pp.\  850--857, 2014.

\bibitem[Singla \& Krause(2013)Singla and Krause]{singla2013truthful}
Singla, Adish and Krause, Andreas.
\newblock Truthful incentives in crowdsourcing tasks using regret minimization
  mechanisms.
\newblock In \emph{WWW}, 2013.

\bibitem[Singla et~al.(2015)Singla, Santoni, Bart{\'o}k, Mukerji, Meenen, and
  Krause]{singla2015incentivizing}
Singla, Adish, Santoni, Marco, Bart{\'o}k, G{\'a}bor, Mukerji, Pratik, Meenen,
  Moritz, and Krause, Andreas.
\newblock Incentivizing users for balancing bike sharing systems.
\newblock In \emph{AAAI}, 2015.

\bibitem[Singla et~al.(2016)Singla, Tschiatschek, and
  Krause]{singla2016actively}
Singla, Adish, Tschiatschek, Sebastian, and Krause, Andreas.
\newblock Actively learning hemimetrics with applications to eliciting user
  preferences.
\newblock In \emph{ICML}, 2016.

\bibitem[Wang et~al.(2016)Wang, Kolar, and Srerbo]{wang2016distributed}
Wang, Jialei, Kolar, Mladen, and Srerbo, Nathan.
\newblock Distributed multi-task learning.
\newblock In \emph{AISTATS}, 2016.

\bibitem[Yan et~al.(2013)Yan, Sundaram, Vishwanathan, and
  Qi]{yan2013distributed}
Yan, Feng, Sundaram, Shreyas, Vishwanathan, SVN, and Qi, Yuan.
\newblock Distributed autonomous online learning: Regrets and intrinsic
  privacy-preserving properties.
\newblock \emph{IEEE Transactions on Knowledge and Data Engineering}, 2013.

\bibitem[Zhou et~al.(2013)Zhou, Liu, Narayan, Ye, Initiative,
  et~al.]{zhou2013modeling}
Zhou, Jiayu, Liu, Jun, Narayan, Vaibhav~A, Ye, Jieping, Initiative, Alzheimer's
  Disease~Neuroimaging, et~al.
\newblock Modeling disease progression via multi-task learning.
\newblock \emph{NeuroImage}, 78:\penalty0 233--248, 2013.

\bibitem[Zinkevich(2003)]{zinkevich2003online}
Zinkevich, Martin.
\newblock Online convex programming and generalized infinitesimal gradient
  ascent.
\newblock In \emph{ICML}, 2003.

\end{thebibliography}
\bibliographystyle{icml2016}
\clearpage

\onecolumn
\appendix
{\allowdisplaybreaks

\section{Outine of the Supplement}
The supplement is composed of the following sections:

\begin{itemize}
	\item Appendix~\ref{appendix_bregman-divergences} introduce properties of the Bregman divergence and additional notation that we require for the later proofs of the regret bounds of the \IOCP and \COCP algorithm.
	\item Appendix~\ref{appendix_propositions} introduces two basic propositions that we need for the proofs in Sections~\ref{appendix_theorems_iocp} and~\ref{appendix_theorems_main}.
	\item Appendix~\ref{appendix_theorems_iocp} provides the proof of the regret bound of the \IOCP algorithm of Theorem~\ref{theorem:IOCP}.
	\item Appendix~\ref{appendix_lemmas} introduces several Lemmas that we  require for the proof of the regret bounds of the \COCP algorithm in Appendices~\ref{appendix_theorems_main} and~\ref{appendix_corollaries_tighter-bounds}.
	\item Appendix~\ref{appendix_idea-weighted-projection} gives the intuitive idea behind using weighted projection, \emph{cf.} Equation~\eqref{eq.weightedproj} in Section~\ref{sec.algorithm}.
	\item Appendix~\ref{appendix_theorems_main} provides the proof of the regret bound of the \COCP algorithm of Theorem~\ref{theorem:COCP}.
	\item Appendix~\ref{appendix_corollaries} provides the proofs of Corollaries~\ref{corollary:COCP.rare} and~\ref{corollary:COCP.special}.
	
	\item Appendix~\ref{appendix_theorems_stochastic} provides the proof of the regret bound of Theorem \ref{theorem:COCP-stochastic}.

	\item Appendix~\ref{appendix_corollaries_tighter-bounds} provides the proof of the tighter regret bound in Corollary~\ref{corollary:COCP.special}.
	\item Appendix~\ref{appendix:experiments} provides details of the user experiment with \names{Airbnb}.
\end{itemize}

%



\section{Preleminaries} \label{appendix_bregman-divergences}
\subsection{Bregman Divergence}
For any strictly convex function $\regularizer^t: \mathbb{R}^d \to \mathbb{R}$, the Bregman divergence $\divergence_{\regularizer^t}$ between $\pmb{a}$, $\pmb{b} \in \mathbb{R}^d$ is defined as the difference between the value of $\regularizer^t$ at $\pmb{a}$, and the first-order Taylor expansion of $\regularizer^t$ around $\pmb{b}$ evaluated at $\pmb{a}$, i.e.
$$\divergence_{\regularizer^t}(\pmb{a}, \pmb{b}) = \regularizer^t(\pmb{a}) - \regularizer^t(\pmb{b}) - \nabla \regularizer^t(\pmb{b}) \cdot (\pmb{a} - \pmb{b}), $$

We use the following properties of the Bregman divergence,  \emph{cf.} \cite{rakhlin2009lecture}:
\begin{itemize}
\item The Bregman divergences is non-negative.
\item The Bregman projection $$\widehat{\pmb{b}} = \argmin_{\pmb{a} \in \solutionspace} \divergence_{\regularizer^t}(\pmb{a}, \pmb{b})$$
onto a convex set $\solutionspace$ exists and is unique.
\item For $\widehat{\pmb{b}}$ defined as in the Bregman projection above and $\pmb{u} \in \solutionspace$, by the generalized Pythagorean theorem,  \emph{cf.} \cite{cesa2006prediction}, the Bregman divergence satisfies
 $$\divergence_{\regularizer^t}(\pmb{u}, \pmb{b}) \geq \divergence_{\regularizer^t}(\pmb{u}, \widehat{\pmb{b}}) + \divergence_{\regularizer^t}(\widehat{\pmb{b}}, \pmb{b}).$$
\item The three-point equality 
$$\divergence_{\regularizer^t}(\pmb{a}, \pmb{b}) + \divergence_{\regularizer^t}(\pmb{b}, \pmb{c}) = \divergence_{\regularizer^t}(\pmb{a}, \pmb{c}) + (\pmb{a} - \pmb{b}) (\nabla \regularizer(\pmb{c}) - \nabla \regularizer(\pmb{b}))$$

follows directly from the definition of the Bregman divergence.
\end{itemize}

\subsection{Notation}
Throughout the supplement we use  $\eta^t_\z = \frac{\eta}{\sqrt{\counter^t_\z}}$ and $\Q^t$ as per Equation~\eqref{eq.qmatrix}. Similar to the definition of $\w^t$ in Section~\ref{subsec.relatedness}, we also define $\x^t$, and $\gradient^t$ as the concatenation of the task specific feature and gradient vectors, \emph{i.e.}
$$\x^t=\big[(\x^t_1)' \ {\cdots} (\x^t_\z)' \ {\cdots} (\x^t_\problems)'\big]'
\qquad
\gradient^t=\big[(\gradient^t_1)' \ {\cdots} (\gradient^t_\z)' \ {\cdots} (\gradient^t_\problems)'\big]'.$$

where for all $t$, $\x^t$ and $\gradient^t$ are $0$ in all positions that do not correspond to task $\z^t$. We also use $\wtilde^{t+1}$ to refer to the concatenation of the updated task specific weights, before any coordination, such that
$$\wtilde^{t+1}=\big[(\w_1)' \ {\cdots} (\w_\z)' \ {\cdots} (\w_\problems)'\big]'.$$

where $\w_z = \w^t_\z$ for $\z \neq z^t$ and $\w	_\z = \wtilde^t_z$ otherwise, \emph{cf.} Algorithm~\ref{algo.COCP} line~\ref{alg1.wold}.



\section{Propositions}\label{appendix_propositions}
In the following we introduce two basic propositions that we need for the proofs in Appendices~\ref{appendix_theorems_iocp} and~\ref{appendix_theorems_main}.

\begin{proposition} \label{proposition:tz}
If $\counter_\z \in \mathbb{R}^+$ for all $\z \in \{1 \dots \problems\}$, and $\sum\limits^\problems_{\z=1} \counter_\z = T ,$
then
$$\sum^\problems_{\z=1} \sqrt{\counter_\z} \leq \sqrt{T \problems} .$$
\end{proposition}

\begin{proof}

Extending and applying the Cauchy-Schwarz inequality, we get
\begin{align*}
\sum^\problems_{\z=1} \sqrt{\counter_\z} &\leq \sqrt{\sum^\problems_{\z=1} 1}\sqrt{\sum^\problems_{\z=1} \counter_\z} \\
&= \sqrt{\problems} \sqrt{T} \\
&= \sqrt{T \problems}
\end{align*}

%
%
%
%
%
\end{proof} 

\begin{proposition} \label{proposition:sumt}
The sum from $\sum^T_{t=1}\frac{1}{\sqrt{t}}$ is bounded by $2 \sqrt{T} - 1$.
\end{proposition}
\begin{proof}
\begin{align*}
\sum^T_{t=1}\frac{1}{\sqrt{t}} &\leq 1 + \int^T_{t=1} \frac{1}{\sqrt{t}} dt \\
&= 1 + \left[2 \sqrt{t} \right]^T_1 \\
&= 2 \sqrt{T} - 1
\end{align*}
\end{proof}



\section{Proof of Theorem~\ref{theorem:IOCP}}
\label{appendix_theorems_iocp}
The regret in Equation~\eqref{eq.regretTotal1}, for any $\wstar \in \solutionspace^*$ can equivalently be written as sum of the regrets of individual learners, such that
\begin{align}
\regret_{\ALG}(T) = \sum_{z = 1}^{\problems} \sum_{t = 1}^{T} \mathds{1}_{\{\z^t = \z\}}\big(\loss^t(\w^t_{\z}) - \loss^t(\wstar_{\z})\big)   \label{eq.regretTotal2}
\end{align}
where $\mathds{1}_{\{\z^t = \z\}}$ is an indicator function to denote the task at time $t$. 

\begin{proof}[Proof of Theorem~\ref{theorem:IOCP}]
Applying Theorem~\ref{theorem:OCP} per individual learner $\OCP_\z \ \forall \z \in [\problems]$, we can state the regret of the algorithm \IOCP using Equation~\eqref{eq.regretTotal2} as follows:
\begin{align*}
\regret_{\IOCP}(T) &\leq \sum_{z = 1}^{\problems} \bigg( \frac{3}{2} \sqrt{\counter^T_\z} \norm{\solutionspace_{max}} \norm{\maxgradient_{max}} \bigg) \\
&= \frac{3}{2} \bigg(\sum_{z = 1}^{\problems} \sqrt{\counter^T_\z}\bigg)  \norm{\solutionspace_{max}} \norm{\maxgradient_{max}} \\
&= \frac{3}{2} \sqrt{T \problems} \norm{\solutionspace_{max}} \norm{\maxgradient_{max}} 
\end{align*}
where the last inequality follows from the Proposition~\ref{proposition:tz}.
\end{proof}



\section{Lemmas}\label{appendix_lemmas}
In this Section we introduce the lemmas that we require for the proof of the regret bounds of the \COCP algorithm in Appendices~\ref{appendix_theorems_main} and~\ref{appendix_corollaries_tighter-bounds}. Applying Lemma~\ref{lemma:linearloss} allows us to replace the loss function with its linearization, similar to \cite{zinkevich2003online}. Lemmas~\ref{lemma:update1} and~\ref{lemma:update2} allow us to get an equivalent update procedure, using the Bregman divergence, and Lemma~\ref{lemma:constraint-minimizer} gives a handle on the linearized regret bound, \emph{cf.} \cite{rakhlin2009lecture}.  Lemma~\ref{lemma:dualitygap} uses the duality gap to upper bound the Bregman divergence between the exact and approximate projection. Lemmas~\ref{lemma:maxdivergence} and~\ref{lemma:rtplusminusrt} provide different upper bounds on the Bregman divergence.

\begin{lemma} \label{lemma:linearloss}
For all $t$ and $\w^t_\z$ there exists a $\gradient^t_\z \in \mathbb{R}^d$ such that  $\loss^t(\w^t_\z)$ can be replaced with $\gradient^t \cdot \w^t_\z$ without loss of generality.
\end{lemma}

\begin{proof}
The loss function affects the regret in two ways: first, the loss function's gradient is used in the update step, and second, the loss function is used to calculate the regret of the algorithm. Let $\gradient^t_\z = \nabla \loss^t(\w^t_\z)$ and consider the linearized loss $\gradient^t_\z \cdot \w^t_\z$. Using the linearized loss, the behavior of the algorithm remains unchanged, since $\nabla \loss^t(\w^t_\z) = \gradient^t_\z$. Further, the regret either increases or remains unchanged, since the loss function is convex, such that for all $\wstar_\z \in \solutionspace_\z$
$$\loss^t(\wstar_\z) \geq \gradient^t_\z \cdot (\wstar_\z - \w^t_\z) + \loss^t(\w^t_\z). $$

Rearranging, we get
\begin{align*}
\loss^t(\w^t_\z) - \loss^t(\wstar_\z) & \leq \gradient^t_\z \cdot \w^t_\z - \gradient^t_\z \cdot \wstar_\z,
\end{align*}

such that using a linearized loss, the regret either remains constant or increases.
\end{proof}

\begin{lemma} \label{lemma:update1}
For $\regularizer^t(\w) = \frac{1}{2} \w' \Q^t \w$, the update rule
$$\wtilde^{t+1} = \w^t - \eta^t_\z \gradient^t$$
is equivalent to the update rule
$$\wtilde^{t+1} = \argmin_{\w \in \mathbb{R}^{d \problems}} \eta \gradient^t \cdot \w + \divergence_{\regularizer^t}(\w, \w^t).$$
\end{lemma}
\begin{proof}
For the second update rule, inserting $\regularizer^t(\w) = \frac{1}{2} \w' \Q^t \w$ into the definition of the Bregman divergence and setting the derivative with respect to $\w$ evaluated at $\wtilde^{t+1}$ to zero, we have
$$\eta \gradient^t + \wtilde^{t+1} \Q^t - \w^t \Q^t = 0$$

Rewriting, using that $\gradient^t$ is non-zero only in entries that correspond to $\z^t$, and applying the definitions of $\Q^t$ and $\eta$, we get
\begin{align*}
\wtilde^{t+1} &= \w^t - \eta \gradient^t (\Q^t)^{-1} \\
&= \w^t - \frac{\eta}{\sqrt{\counter^t_\z}} \gradient^t \\
&= \w^t - \eta^t_\z \gradient^t.
\end{align*}
\end{proof}

\begin{lemma} \label{lemma:update2}
For $\regularizer^t(\w) = \frac{1}{2} \w' \Q^t \w$, the update rule
$$\w^{t+1} = \argmin_{\w \in \solutionspace^*} \divergence_{\regularizer^t}(\w, \wtilde^{t+1}),$$
where $ \wtilde^{t+1} = \w^t - \eta^t_\z \gradient^t$,
is equivalent to the update rule
$$\w^{t+1} = \argmin_{\w \in \solutionspace^*} \eta \gradient^t \cdot \w + \divergence_{\regularizer^t}(\w, \w^t)$$
\end{lemma}
\begin{proof}

Applying the definition of $\regularizer^t(\w)$, we can rewrite 
\begin{align*}
\w^{t+1} &= \argmin_{\w \in \solutionspace^*} \divergence_{\regularizer^t}(\w, \wtilde^{t+1}) \\
&= \argmin_{\w \in \solutionspace^*} \frac{1}{2} (\w - \w^t + \eta \gradient^t (\Q^t)^{-1})' \Q^t (\w - \w^t + \eta \gradient^t (\Q^t)^{-1}) \\
&= \argmin_{\w \in \solutionspace^*} \eta \gradient^t \cdot \w  + \frac{1}{2} (\w - \w^t)' \Q^t (\w - \w^t) \\
&= \argmin_{\w \in \solutionspace^*} \eta \gradient^t \cdot \w + \divergence_{\regularizer^t}(\w, \w^t)
\end{align*}
\end{proof}

\begin{lemma} \label{lemma:constraint-minimizer}
If $\w^{t+1}$ is the constraint minimizer of the objective $\eta \gradient^t \cdot \w + \divergence_{\regularizer^t}(\w, \w^t)$ as stated in Lemma \ref{lemma:update2}, then for any \pmb{a} in the solution space,
$$\eta \gradient^t \cdot (\w^{t+1} - \pmb{a}) \leq \divergence_{\regularizer^t}(\pmb{a}, \w^t) - \divergence_{\regularizer^t}(\pmb{a}, \w^{t+1}) - \divergence_{\regularizer^t}(\w^{t+1}, \w^t). $$
\end{lemma}
\begin{proof}
Since $\w^{t+1}$ is the constraint minimizer of the objective $\eta \gradient^t \cdot \w + \divergence_{\regularizer^t}(\w, \w^t)$, any vector pointing away from $\w^{t+1}$ into the solution space has a positive product with the gradient of the objective at $\w^{t+1}$, such that
$$ 0 \leq (\pmb{a} - \w^{t+1}) \cdot (\eta \gradient^t + \nabla \regularizer^t(\w^{t+1}) -\nabla \regularizer^t(\w^t)). $$
Rewriting and using the three-point equality, we get
\begin{align*}
\eta \gradient^t \cdot (\w^{t+1} - \pmb{a}) &\leq (\pmb{a} - \w^{t+1}) \cdot (\nabla \regularizer^t(\w^{t+1}) - \nabla \regularizer^t(\w^t)) \\
&= \divergence_{\regularizer^t}(\pmb{a}, \w^t) - \divergence_{\regularizer^t}(\pmb{a}, \w^{t+1}) - \divergence_{\regularizer^t}(\w^{t+1}, \w^t).
\end{align*}
\end{proof}

\begin{lemma} \label{lemma:dualitygap}
If $\what^{t+1}$ is the exact solution of 
$$\argmin_{\w \in \solutionspace^*} \divergence_{\regularizer^t}(\w, \wtilde^{t+1})$$
and $\w^{t+1} \in \solutionspace^*$ is an approximate solution with duality gap less than $\maxdualitygap$, then
$$\maxdualitygap \geq \divergence_{\regularizer^t}(\what^{t+1}, \w^{t+1}) .$$
\end{lemma}
\begin{proof}
The duality gap is defined as the difference between the primal and dual value of the solution. The dual value is upper bounded by the optimal solution and thus less than or equal to $\divergence_{\regularizer^t}(\what^{t+1}, \wtilde^{t+1})$. Thus, for the primal solution $\divergence_{\regularizer^t}(\w^{t+1}, \wtilde^{t+1})$ with duality gap less than $\maxdualitygap$, we have
$$\maxdualitygap \geq \divergence_{\regularizer^t}(\w^{t+1}, \wtilde^{t+1}) - \divergence_{\regularizer^t}(\what^{t+1}, \wtilde^{t+1}) $$
Note that $\what^{t+1}$ is the projection of $\wtilde^{t+1}$ onto $\solutionspace^*$ and $\w^{t+1} \in \solutionspace^*$. Thus, using the propertiesof the Bregman divergence we can apply the generalized Pythagorean theorem such that
$$\divergence_{\regularizer^t}(\w^{t+1}, \wtilde^{t+1}) \geq \divergence_{\regularizer^t}(\w^{t+1}, \what^{t+1}) + \divergence_{\regularizer^t}(\what^{t+1}, \wtilde^{t+1}) $$

Inserting into the above inequality we get the result.
\end{proof}

\begin{lemma} \label{lemma:maxdivergence}
For $\regularizer^t(\w) = \frac{1}{2} \w' \Q^t \w$ and $\pmb{a}$ and $\pmb{b} \in \solutionspace$,
$$\divergence_{\regularizer^t}(\pmb{a}, \pmb{b}) \leq \frac{1}{2} \norm{\solutionspace_{max}}^2 \sqrt{t \problems}  $$
\end{lemma}

\begin{proof}
Using the definition of $\Q^t$, noting that $\norm{\pmb{a}_\z - \pmb{b}_\z}^2 \leq \norm{\solutionspace_{max}}^2$, and applying Proposition~\ref{proposition:tz} we can write
\begin{align*}
\divergence_{\regularizer^t}(\pmb{a}, \pmb{b}) &= \frac{1}{2} (\pmb{a} - \pmb{b})' \Q^t (\pmb{a} - \pmb{b}) \\
&= \frac{1}{2} \sum^\problems_{\z=1} \norm{\pmb{a}_\z - \pmb{b}_\z}^2 \sqrt{\counter^t_\z} \\
&\leq \frac{1}{2} \norm{\solutionspace_{max}}^2 \sum^\problems_{\z=1} \sqrt{\counter^t_\z} \\
&\leq \frac{1}{2} \norm{\solutionspace_{max}}^2 \sqrt{t \problems} 
\end{align*}
\end{proof}

\begin{lemma} \label{lemma:rtplusminusrt}
For any two $\pmb{a}^t$, $\pmb{b}^t \in \solutionspace$,
$$\sum^T_{t=1} \divergence_{\regularizer^{t+1}}(\pmb{a}^t, \pmb{b}^t) - \divergence_{\regularizer^t}(\pmb{a}^t, \pmb{b}^t) \leq \frac{1}{2} \norm{\solutionspace_{max}}^2 \sqrt{T \problems} .$$
\end{lemma}
\begin{proof}
Applying our definition of $\regularizer^t$, we can rewrite 
$$ \sum^T_{t=1} \divergence_{\regularizer^{t+1}}(\pmb{a}^t, \pmb{b}^t) - \divergence_{\regularizer^t}(\pmb{a}^t, \pmb{b}^t) = \frac{1}{2} \sum^T_{t=1} (\pmb{a}^t - \pmb{b}^t)' (\Q^{t+1} - \Q^t) (\pmb{a}^t - \pmb{b}^t). $$

Note that 
$$ (\Q^{t+1} - \Q^t) =
\begin{bmatrix}
\sqrt{\counter^{t+1}_1} - \sqrt{\counter^t_1} & & 0  \\
& \ddots & \\
0 & & \sqrt{\counter^{t+1}_\problems} - \sqrt{\counter^t_\problems}
\end{bmatrix}. $$

Applying Proposition~\ref{proposition:tz} and using $\norm{\pmb{a}_\z - \pmb{b}_\z}^2 \leq \norm{\solutionspace_\z}^2 \leq \norm{\solutionspace_{max}}^2$  we get
\allowdisplaybreaks
\begin{align*}
\sum^T_{t=1} \divergence_{\regularizer^{t+1}}\norm{\pmb{a}^t, \pmb{b}^t} - \divergence_{\regularizer^t}(\pmb{a}^t, \pmb{b}^t) &= \frac{1}{2}  \sum^T_{t=1} \sum^\problems_{\z=1} \left(\sqrt{\counter^{t+1}_\z} - \sqrt{\counter^t_\z}\right) \norm{\pmb{a}^t_\z - \pmb{b}^t_\z}^2 \\
&\leq \frac{1}{2} \norm{\solutionspace_{max}}^2 \sum^\problems_{\z=1} \sum^T_{t=1} \left(\sqrt{\counter^{t+1}_\z} - \sqrt{\counter^t_\z} \right) \\
&= \frac{1}{2} \norm{\solutionspace_{max}}^2  \sum^\problems_{\z=1} \left(\sqrt{\counter^{T+1}_\z} - \sqrt{\counter^{1}_\z} \right) \\
&= \frac{1}{2} \norm{\solutionspace_{max}}^2  \left( \sum^\problems_{\z=1} \left(\sqrt{\counter^{T+1}_\z} \right) - 1  \right) \\
&= \frac{1}{2} \norm{\solutionspace_{max}}^2 \left( \sum^\problems_{\z=1} \left(\sqrt{\counter^T_\z} \right) + \sqrt{\counter^{T+1}_{\z^{T+1}}} - \sqrt{\counter^T_{\z^{T+1}}}  - 1  \right)\\
&= \frac{1}{2} \norm{\solutionspace_{max}}^2  \left( \sum^\problems_{\z=1} \left(\sqrt{\counter^T_\z} \right) + \sqrt{\counter^T_{\z^{T+1}} + 1} - \sqrt{\counter^T_{\z^{T+1}}}  - 1  \right) \\
&\leq \frac{1}{2} \norm{\solutionspace_{max}}^2  \sum^\problems_{\z=1} \sqrt{\counter^T_\z} \\
&\leq \frac{1}{2} \norm{\solutionspace_{max}}^2 \sqrt{T \problems} \\
\end{align*}
\end{proof}


\section{Idea of Weighted Projection for \COCP}\label{appendix_idea-weighted-projection}
The update in Algorithm~\ref{algo.OCP} line~\ref{alg3.update} can be equivalently written as 
$$\wtilde^{t+1} = \w^t - \eta^t_\z \gradient^t.$$

As shown in Lemma~\ref{lemma:update1}, we can rewrite this as
$$\wtilde^{t+1} = \argmin_{\w \in \mathbb{R}^{d \problems}} \eta \gradient^t \cdot \w + \divergence_{\regularizer^t}(\w, \w^t),$$
using the regularizer $\regularizer^t(\w) = \frac{1}{2} \w \cdot \Q^t \w$.

Intuitively, the central coordinator \COCP restricts the solution to $\solutionspace^*$, such that the update after coordination can be rewritten as 
\begin{align*}
\w^{t+1} &= \argmin_{\w \in \solutionspace^*} \eta \gradient^t \cdot \w + \divergence_{\regularizer^t}(\w, \w^t) \\
&= \argmin_{\w \in \solutionspace^*} \eta \gradient^t \cdot \w  + \frac{1}{2} (\w - \w^t) \cdot \Q^t (\w - \w^t) \\
&= \argmin_{\w \in \solutionspace^*} \frac{1}{2} (\w - \w^t + \eta \gradient^t (\Q^t)^{-1}) \cdot \Q^t (\w - \w^t + \eta \gradient^t (\Q^t)^{-1}) \\
&= \argmin_{\w \in \solutionspace^*} \frac{1}{2} (\w - \wtilde^{t+1}) \cdot \Q^t (\w - \wtilde^{t+1}),
\end{align*}

which is equal to the weighted projection introduced in Equation~\eqref{eq.weightedproj}. Using weights defined by some other heuristics could in general lead to a higher regret. For instance, in Figure~\ref{fig.single} we show the increase in regret of the \COCP algorithm, when setting $\Q^t$ as the identity matrix.

%
%
%
%
%
%
%
%
%
%



\section{Proof of Theorem~\ref{theorem:COCP}}\label{appendix_theorems_main}
In the following we provide the proof of Theorem~\ref{theorem:COCP}, using notation and results of the earlier sections of the supplement. Unlike earlier work (\emph{e.g.} \cite{zinkevich2003online, rakhlin2009lecture}), in our setting projections are allowed to be noisy and therefore, the solution may not be a constraint minimizer of the projection. Additionally, in our setting coordination may occur only sporadically, and thus intermediary solutions may not be in $\solutionspace^*$. To keep track of whether coordination occurred, we define indicator functions and handle the special case of coordination at time $t$ without coordination at time $t-1$ separately. 


\begin{proof}{Proof of Theorem~\ref{theorem:COCP}}
\subsection*{Preparation}
We define $\what^t$ as the exact solution of the projection onto $\solutionspace^*$, such that
$$\what^t = \argmin_{\w \in \solutionspace^*} (\w - \wtilde^{t+1})' \Q^t (\w - \wtilde^{t+1}) .$$

Recall that $\xi^t$ is $1$ with probability $\alpha$ and $0$ with probability $(1 - \alpha)$. The algorithm projects onto $\solutionspace^*$ if $\xi^t = 1$ and onto $\solutionspace_\z$ if $\xi^t = 0$. We define the indicator functions
\begin{align*}
\inside = 
\left\{
	\begin{array}{ll}
		1 & \mbox{if } \xi^t = 1\\
		0 & \text{otherwise} .
	\end{array}
\right.
\end{align*}
and the inverse

\begin{align*}
\outside = 
\left\{
	\begin{array}{ll}
		1 & \mbox{if } \xi^t = 0\\
		0 & \text{otherwise} .
	\end{array}
\right.
\end{align*}

as well as
\begin{align*}
\bad = 
\left\{
	\begin{array}{ll}
		1 & \mbox{if } \xi^{t-1} = 0 \text{ and } \xi^t = 1 \\
		0 & \text{otherwise} .
	\end{array}
\right.
\end{align*}
and the inverse
\begin{align*}
\good = 
\left\{
	\begin{array}{ll}
		1 & \mbox{if } \xi^{t-1} = 1 \text{ or } \xi^t = 0 \\
		0 & \text{otherwise} .
	\end{array}
\right.
\end{align*}

Our goal is to upper bound the regret, which, using Lemma~\ref{lemma:linearloss}, we can write as $$ \regret_{\COCP}(T)  = \sum^T_{t=1} \gradient^t_\zt \cdot (\w^t_\zt - \wstar_\zt)  .$$ Using the definitions above, we rewrite
\begin{align*}
\sum^T_{t=1} \gradient^t_\zt \cdot (\w^t_\zt - \wstar_\zt) &= \sum^T_{t=1} \gradient^t \cdot (\w^t - \wstar) \\
&= \sum^T_{t=1} \gradient^t \cdot (\what^{t+1} - \wstar) + \sum^T_{t=1} \gradient^t \cdot (\w^t - \what^{t+1}) \\
&= \sum^T_{t=1} \inside \gradient^t \cdot (\what^{t+1} - \wstar) + \sum^T_{t=1} \outside \gradient^t \cdot (\what^{t+1} - \wstar) \\
&\phantom{=} + \sum^T_{t=1} \good \gradient^t \cdot (\w^t - \what^{t+1}) + \sum^T_{t=1} \bad \gradient^t \cdot (\w^t - \what^{t+1}) .
\end{align*}

and further upper bound each sum individually.

Throughout the proof, we use the Bregman divergence with the regularizer $\regularizer^t(\w) = \frac{1}{2} \w' \Q^t \w$, and apply Lemmas~\ref{lemma:update1} and~\ref{lemma:update2} to get an equivalent update procedure.

\subsection*{Step 1: First sum}
Applying Lemma~\ref{lemma:constraint-minimizer} with $\what^{t+1}$ as the constraint minimizer of the objective $\eta \gradient^t \cdot \w + \divergence_{\regularizer^t}(\w, \w^t)$ and $\wstar \in \solutionspace^*$, we have
$$ \eta \gradient^t \cdot (\what^{t+1} - \wstar) \leq \divergence_{\regularizer^t}(\wstar, \w^t) - \divergence_{\regularizer^t}(\wstar, \what^{t+1}) - \divergence_{\regularizer^t}(\what^{t+1}, \w^t) . $$

Adding over time,
\begin{align*}
\eta \sum^T_{t=1} \inside \gradient^t \cdot (\what^{t+1} - \wstar) &\leq \sum^T_{t=1} \inside \left( \divergence_{\regularizer^t}(\wstar, \w^t) - \divergence_{\regularizer^t}(\wstar, \what^{t+1}) - \divergence_{\regularizer^t}(\what^{t+1}, \w^t) \right) \\
&= \sum^T_{t=1} \left(\insidet{t+1} \divergence_{\regularizer^{t+1}}(\wstar, \w^{t+1}) - \inside \divergence_{\regularizer^t}(\wstar, \what^{t+1}) - \inside \divergence_{\regularizer^t}(\what^{t+1}, \w^t) \right) \\
& \phantom{\leq} + \insidet{1} \divergence_{\regularizer^1}(\wstar, \w^1) - \insidet{T+1}  \divergence_{\regularizer^{T+1}}(\wstar, \w^{T+1}) \\
&\leq \sum^T_{t=1} \left(\insidet{t+1} \divergence_{\regularizer^{t+1}}(\wstar, \w^{t+1}) - \inside \divergence_{\regularizer^t}(\wstar, \w^{t+1}) \right) \\
& \phantom{=} + \sum^T_{t=1} \inside \left( \divergence_{\regularizer^t}(\wstar, \w^{t+1}) - \divergence_{\regularizer^t}(\wstar, \what^{t+1}) \right) \\
& \phantom{=} + \insidet{1} \divergence_{\regularizer^1}(\wstar, \w^1) .
\end{align*}

In the following we upper bound each term individually. For now we leave the first term  unchanged and provide an upper bound in step 3 by combining it with the results of step 2.

For the second term, we use that for our choice of $\regularizer$, the square root of the Bregman divergence is a norm and therefore satisfies the triangle inequality. Thus,
$$ \sqrt{\divergence_{\regularizer^t}(\wstar, \w^{t+1})} \leq \sqrt{\divergence_{\regularizer^t}(\wstar, \what^{t+1})} + \sqrt{\divergence_{\regularizer^t}(\what^{t+1}, \w^{t+1})} .$$

Squaring both sides, we have
$$\divergence_{\regularizer^t}(\wstar, \w^{t+1}) \leq \divergence_{\regularizer^t}(\wstar, \what^{t+1}) + \divergence_{\regularizer^t}(\what^{t+1}, \w^{t+1}) + 2 \sqrt{\divergence_{\regularizer^t}(\what^{t+1}, \w^{t+1}) \divergence_{\regularizer^t}(\wstar, \what^{t+1})} .$$

Applying Lemmas~\ref{lemma:dualitygap} and~\ref{lemma:maxdivergence}, we get
$$ \divergence_{\regularizer^t}(\wstar, \w^{t+1}) - \divergence_{\regularizer^t}(\wstar, \what^{t+1}) \leq \maxdualitygap + \sqrt{2 \maxdualitygap} (t\problems)^{1/4} \norm{\solutionspace_{max}} .$$

For the third term, using that $\Q^1$ is 1 in exactly one position, we have
$$\insidet{1} \divergence_{\regularizer^1}(\wstar, \w^1) \leq \insidet{1} \frac{1}{2} \norm{\solutionspace_{max}}^2 .$$ 

Combining and dividing by $\eta$, we get the upper bound for the first sum
\begin{align*}
\sum^T_{t=1} \inside \gradient^t \cdot (\what^{t+1} - \wstar) &\leq \frac{1}{\eta} \sum^T_{t=1} \insidet{t+1} \divergence_{\regularizer^{t+1}}(\wstar, \w^{t+1}) - \inside \divergence_{\regularizer^t}(\wstar, \w^{t+1}) \\
&\phantom{\leq} + \frac{1}{\eta} \sum^T_{t=1}  \inside \left( \maxdualitygap + \sqrt{2 \maxdualitygap}(t\problems)^{1/4} \norm{\solutionspace_{max}} \right) \\
&\phantom{\leq} + \frac{1}{2 \eta} \insidet{1} \norm{\solutionspace_{max}}^2.
\end{align*}

\subsection*{Step 2: Second sum}
Similar to step 1, we get
\begin{align*}
\eta \sum^T_{t=1} \outside \gradient^t \cdot (\what^{t+1} - \wstar) &\leq \sum^T_{t=1} \left( \outsidet{t+1} \divergence_{\regularizer^{t+1}}(\wstar, \w^{t+1}) -  \outside \divergence_{\regularizer^t}(\wstar, \w^{t+1}) \right) \\
& \phantom{=} + \sum^T_{t=1} \outside \left( \divergence_{\regularizer^t}(\wstar, \w^{t+1}) - \divergence_{\regularizer^t}(\wstar, \what^{t+1}) \right) \\
& \phantom{=} + \outsidet{1} \divergence_{\regularizer^1}(\wstar, \w^1) .
\end{align*}

As in step 1, we leave the first term unchanged. For the second term, note that $w^{t+1}$ is not project onto $\solutionspace^*$, and thus $\what^t = \w^t$ for all $t$, such that 
$$ \sum^T_{t=1} \outside \left( \divergence_{\regularizer^t}(\wstar, \w^{t+1}) - \divergence_{\regularizer^t}(\wstar, \what^{t+1}) \right) = 0 .$$

For the third term, similar to step 1, we have 
$$\outsidet{1} \divergence_{\regularizer^1}(\wstar, \w^1) \leq \outsidet{1} \frac{1}{2} \norm{\solutionspace_{max}}^2 .$$

Combining, we get the upper bound for the second sum
\begin{align*}
\sum^T_{t=1} \outside \gradient^t \cdot (\what^{t+1} - \wstar) &\leq \frac{1}{\eta} \sum^T_{t=1} \outsidet{t+1} \divergence_{\regularizer^{t+1}}(\wstar, \w^{t+1}) - \outside \divergence_{\regularizer^t}(\wstar, \w^{t+1}) \\
&\phantom{=} + \frac{1}{2 \eta} \outsidet{1} \norm{\solutionspace_{max}}^2.
\end{align*}

\subsection*{Step 3: Combination of steps 1 and 2}
Note that $\inside + \outside = 1$ for all t. Thus, the first terms of step 1 and 2 sum to
$$\frac{1}{\eta} \sum^T_{t=1} \divergence_{\regularizer^{t+1}}(\wstar, \w^{t+1}) - \divergence_{\regularizer^t}(\wstar, \w^{t+1}) .$$

Using Lemma~\ref{lemma:rtplusminusrt}, we get
$$\frac{1}{\eta} \sum^T_{t=1} \divergence_{\regularizer^{t+1}}(\wstar, \w^{t+1}) - \divergence_{\regularizer^t}(\wstar, \w^{t+1}) \leq \frac{1}{2 \eta} \norm{\solutionspace_{max}}^2 \sqrt{T \problems} .$$

Summing the remaining terms and again noting that $\inside + \outside = 1$, we get the upper bound for the first and second sum
\begin{align*}
\sum^T_{t=1} \gradient^t \cdot (\what^{t+1} - \wstar) &\leq \frac{1}{2 \eta} \norm{\solutionspace_{max}}^2 \sqrt{T \problems} + \frac{1}{2 \eta} \norm{\solutionspace_{max}}^2 \\
&\phantom{\leq} + \frac{1}{\eta} \sum^T_{t=1}  \inside \left( \maxdualitygap + \sqrt{2 \maxdualitygap}(t\problems)^{1/4} \norm{\solutionspace_{max}} \right).
\end{align*}

\subsection*{Step 4: Third sum}

To upper bound $\sum^T_{t=1} \good \gradient^t \cdot (\w^t - \what^{t+1})$ we start by using H{\"o}lder's inequality (see for example \cite{beckenbach2012inequalities}) to get
$$\gradient^t \cdot (\w^t - \w^{that+1})  \leq \norm{\gradient^t}^*_{\Q^t} \norm{\w^t - \what^{t+1}}_{\Q^t},$$
where
$$ \norm{\gradient^t}^*_{\Q^t} = \max_{\pmb{x}} \pmb{x} \cdot \gradient^t : \norm{\pmb{x}}_{\Q^t} \leq 1 .$$

For the norm $\norm{\w^t - \what^{t+1}}_{\Q^t}$ we applya Lemma~\ref{lemma:constraint-minimizer} with $\what^{t+1}$ as the constraint minimizer of the objective $\eta \gradient^t \cdot \w + \divergence_{\regularizer^t}(\w, \w^t)$ with $\w^t \in \solutionspace^*$. Using the symmetry of the Bregman divergence for our choice of $\regularizer^t$,
$$ \eta \gradient^t \cdot (\what^{t+1} - \w^t) \leq - 2 \divergence_{\regularizer^t}(\w^{t+1}, \w^t) $$
and thus

$$ \divergence_{\regularizer^t}(\what^{t+1}, \w^t) \leq \frac{1}{2} \eta \gradient^t \cdot (\w^t - \what^{t+1}) .$$

Note that $\divergence_{\regularizer^t}(\w^t, \what^{t+1}) = \frac{1}{2} \norm{\w^t - \what^{t+1}}^2_{\Q^t}$ and thus, 
$$ \norm{\w^t - \what^{t+1}}^2_{\Q^t} \leq \eta \gradient^t \cdot (\w^t - \what^{t+1}) .$$

Using H{\"o}lder's inequality on the right side of the inequality, we get
$$\norm{\w^t - \w^{t+1}}_{\Q^t} \leq \eta \norm{\gradient^t}^*_{\Q^t}.$$
Therefore,
$$ \gradient^t \cdot (\w^t - \w^{t+1}) \leq \eta (\norm{\gradient^t}^*_{\Q^t})^2 $$

We now apply the definition of the dual norm to rewrite $\norm{\gradient^t}^*_{\Q^t}$. Note that $\gradient^t$ is non-zero only in position $\z^t$ and thus
\begin{align*}
\norm{\gradient^t}^*_{\Q^t} &= \max_{\pmb{x}} \pmb{x} \cdot \gradient^t : \norm{\pmb{x}}_{\Q^t} \leq 1 \\
&= \max_{\pmb{x}_{\z^t}} \pmb{x}_{\z^t} \gradient^t_{\z^t} : \left(\norm{\pmb{x}_{\z^t}}^2_2 \sqrt{\counter^t_{\z^t}}\right)^{1/2} \leq 1 \\
&= \max_{\pmb{x}_{\z^t}} \pmb{x}_\z \gradient^t_{\z^t} : \norm{\pmb{x}_{\z^t}}_2 \leq \frac{1}{\sqrt{\counter^t_{\z^t}}^{1/2}}  \\
&\leq \max_{\pmb{x}_{\z^t}} \norm{\pmb{x}_{\z^t}}_2 \norm{\gradient^t_{\z^t}}_2 : \norm{\pmb{x}_{\z^t}}_2 \leq \frac{1}{\sqrt{\counter^t_{\z^t}}^{1/2}}  \\
\end{align*}
The maximum is achieved at $\norm{\pmb{x}_{\z^t}}_2 = \frac{1}{\sqrt{\counter^t_{\z^t}}^{1/2}}$. Thus,
$$ \norm{\gradient^t}^*_{\Q^t} \leq \norm{\gradient^t_{\z^t}}_2 \frac{1}{\sqrt{\counter^t_{\z^t}}^{1/2}} .$$

Inserting, summing, and using Propositions~\ref{proposition:tz} and~\ref{proposition:sumt}, we get the upper bound for the third sum,

\begin{align*}
\sum^T_{t=1}  \good  \gradient^t \cdot (\w^t - \w^{t+1}) &\leq \eta \sum^T_{t=1}  \sum^\problems_{\z=1} \norm{\gradient^t_\z}^2_2 \frac{1}{\sqrt{\counter^t_\z}} \\
&\leq \eta \sum^\problems_{\z=1}  \sum^T_{t=1} \norm{\maxgradient_\z}^2 \frac{1}{\sqrt{\counter^t_\z}}  \\
&\leq 2 \eta \sum^\problems_{\z=1} \norm{\maxgradient_\z}^2  \left( \sqrt{\counter^T_\z} -1 \right) \\
&\leq 2 \eta \norm{\maxgradient_{max}}^2 \sum^\problems_{\z=1} \left( \sqrt{\counter^T_\z} -1 \right) \\
&\leq 2 \eta \norm{\maxgradient_{max}}^2 \sqrt{T\problems} - 2 \eta \norm{\maxgradient_{max}}^2 \problems
\end{align*}

\subsection*{Step 5: Fourth sum}
For the fourth sum we use that 
$$ \gradient^t \cdot (\w^t - \what^{t+1}) = \gradient^t_\zt \cdot (\w^t_\zt - \what^{t+1}_\zt).$$

Using the Cauchy-Schwarz inequality, we get
\begin{align*}
\gradient^t_\zt \cdot (\w^t_\zt - \what^{t+1}_\zt) &\leq \norm{\gradient^t_\zt}_2 \norm{\w^t_\zt - \what^{t+1}_\zt}_2 \\
&\leq \norm{\solutionspace_{max}} \norm{\maxgradient_{max}}
\end{align*}

Thus,
$$\sum^T_{t=1}  \bad \gradient^t \cdot (\w^t - \what^{t+1}) \leq \sum^T_{t=1}  \bad \norm{\solutionspace_{max}} \norm{\maxgradient_{max}}.$$

\subsection*{Step 6: Combination}
Adding the results from steps 1 to 5, we get the result
\begin{align*}
\sum^T_{t=1} \gradient^t \cdot (\w^t - \wstar) &\leq \frac{1}{2 \eta} \norm{\solutionspace_{max}}^2 \sqrt{T \problems} +  2 \eta \norm{\maxgradient_{max}}^2 \sqrt{T\problems} \\
&\phantom{\leq} +\sum^T_{t=1}  \bad \norm{\solutionspace_{max}} \norm{\maxgradient_{max}}\\
&\phantom{\leq} + \frac{1}{\eta} \sum^T_{t=1}  \inside \left( \maxdualitygap + \sqrt{2 \maxdualitygap}(t\problems)^{1/4} \norm{\solutionspace_{max}} \right) \\
&\phantom{\leq} + \frac{1}{2 \eta} \norm{\solutionspace_{max}}^2  - 2 \eta \norm{\maxgradient_{max}}^2 \problems.
\end{align*}
\end{proof}


\section{Proof of Corollaries}\label{appendix_corollaries}
By plugging in specific algorithmic parameters into Theorem~\ref{theorem:COCP} we can get more concrete regret bounds on the \COCP algorithm. In the two Corollaries~\ref{corollary:COCP.rare} and~\ref{corollary:COCP.special} we provide no-regret bounds for two common parametric choices, and note that similar no-regret bounds can also be achieved for different parameters.


\subsection{Proof of Corollary~\ref{corollary:COCP.rare}}
\begin{proof}[Proof of Corollary~\ref{corollary:COCP.rare}]

Inserting $\eta = \frac{1}{2} \frac{\norm{\solutionspace_{max}}}{\norm{\maxgradient_{max}}}$ into the results of Theorem~\ref{theorem:COCP}, taking the expected value over $\xi^t$, and using that $\problems \geq 1$, we get
\begin{align*}
\expectedvalue{\sum^T_{t=1} \gradient^t \cdot (\w^t - \wstar)} &\leq 2 \sqrt{T \problems} \norm{\solutionspace_{max}} \norm{\maxgradient_{max}} \\
&\phantom{\leq} + \alpha (1 - \alpha) T \norm{\solutionspace_{max}}  \norm{\maxgradient_{max}} \\
&\phantom{\leq} + 2 \alpha \frac{\norm{\maxgradient_{max}}}{\norm{\solutionspace_{max}}}  \sum^T_{t=1}   \left( \maxdualitygap + \sqrt{2 \maxdualitygap}(t\problems)^{1/4} \norm{\solutionspace_{max}} \right) .
\end{align*}

For the third term, sing $\maxdualitygap = \cbeta (1 - \beta)^2 \frac{\sqrt{\problems}}{\sqrt{t}} \norm{\solutionspace_{max}}^2$ where $\cbeta \geq 0$, $\beta \in [0,1]$, then
\begin{align*}
\maxdualitygap + \sqrt{2 \maxdualitygap}(t\problems)^{1/4} \norm{\solutionspace_{max}} &= \cbeta (1 - \beta)^2 \frac{\sqrt{\problems}}{\sqrt{t}} \norm{\solutionspace_{max}}^2 + (1 - \beta) \sqrt{2 \cbeta} \sqrt{\problems}  \norm{\solutionspace_{max}}^2 \\
&\leq \cbeta (1 - \beta) \sqrt{\problems} \norm{\solutionspace_{max}}^2 + (1 - \beta) \sqrt{2 \cbeta} \sqrt{\problems}   \norm{\solutionspace_{max}}^2 \\
&= (1 - \beta) \sqrt{\problems} (\cbeta + \sqrt{2 \cbeta}) \norm{\solutionspace_{max}}^2 ,
\end{align*}

and, using Proposition \ref{proposition:sumt} for the sum,
$$\sum^T_{t=1} \maxdualitygap +  \sqrt{2 \maxdualitygap}(t\problems)^{1/4} \norm{\solutionspace_{max}}  \leq \alpha (1 - \beta) T \sqrt{\problems} (\cbeta + \sqrt{2 \cbeta}) \norm{\solutionspace_{max}}^2 .$$

Inserting $\alpha = \frac{\calpha}{\sqrt{T}}$  where $\calpha \in [0, \sqrt{T}]$, we get
\begin{align*}
\expectedvalue{\sum^T_{t=1} \gradient^t \cdot (\w^t - \wstar)} &\leq 2 \sqrt{T \problems} \norm{\solutionspace_{max}}  \norm{\maxgradient_{max}} \\
&\phantom{\leq}  + \sqrt{T} \calpha \left(1 - \frac{\calpha}{\sqrt{T}} \right) \norm{\solutionspace_{max}}  \norm{\maxgradient_{max}} \\
&\phantom{\leq} + 2 \sqrt{T \problems} \calpha (\cbeta + \sqrt{2 \cbeta}) (1 - \beta) \norm{\solutionspace_{max}}  \norm{\maxgradient_{max}} .
\end{align*}

\end{proof}


\subsection{Proof of Corollary~\ref{corollary:COCP.special}}
\begin{proof}[Proof of Corollary~\ref{corollary:COCP.special}]

Inserting $\eta = \frac{1}{2} \frac{\norm{\solutionspace_{max}}}{\norm{\maxgradient_{max}}}$ into the results of Theorem~\ref{theorem:COCP}, we note that the first and fourth term are identical to the proof of Corollary~\ref{corollary:COCP.rare}. 

For $\xi^t = 1$ and $\dualitygap^t = 0  \ \forall t \in [T]$, the second and third term equals zero.
Thus, we get 
$$\sum^T_{t=1} \gradient^t \cdot (\w^t - \wstar) \leq 2 \sqrt{T \problems} \norm{\solutionspace_{max}}  \norm{\maxgradient_{max}} .$$

A more careful analysis yields the tighter regret bound
$$\sum^T_{t=1} \gradient^t \cdot (\w^t - \wstar) \leq \frac{3}{2} \sqrt{T \problems} \norm{\solutionspace_{max}}  \norm{\maxgradient_{max}}$$

The proof for the tighter bound is a bit more involved and provided in Appendix~\ref{appendix_corollaries_tighter-bounds}.

\end{proof}

\section{Proof of Theorem \ref{theorem:COCP-stochastic}}
\label{appendix_theorems_stochastic}
In this section, we provide the proof of Theorem~\ref{theorem:COCP-stochastic}, showing the improved bounds of the \COCP algorithm in a simple $B$-batch setting, in which a task instance is repeated $B$ times before choosing a new one. In the setting considered in this theorem, we have $d=1$ with shared parameter structure (\emph{cf.} Section~\ref{subsec.relatedness}) and a $\epsilon$-insensitive loss function given by $\loss^t(\w^t_{\z^t}) = 0  \textnormal{ if } \lvert \w^t_{\z^t} - c^* \rvert \leq \epsilon$, else  $\loss^t(\w^t_{\z^t}) = \lvert \w^t_{\z^t} - c^* \rvert - \epsilon$, where $\epsilon > 0$ and $c^* \in \mathbb{R}$ is a constant.
%

\begin{proof}[Proof of Theorem~\ref{theorem:COCP-stochastic}]
We denote the task observed in the first batch by $z$, such that $z^t = z$ for $t \in 1 \ldots B$. 

\subsection*{Step 1}
In the $B$-batch setting, the learner $\OCP_z$, corresponding to the first task receives the first $B$ task instances. Our key observation is that at the end of this batch, after $B$ time steps, where $B \geq \big\lceil(\frac{\norm{\solutionspace_{max}}}{\epsilon} + \frac{1}{2})^2\big\rceil$, the weight vector of this learner satisfies the condition $\lvert \w^t_{\z} - c^* \rvert \leq \epsilon$. 

First, using $\eta = \frac{\norm{\solutionspace_{max}}}{\norm{\maxgradient_{max}}}$ and $\norm{\maxgradient_{max}} = \norm{\gradient^t_{\z^t}}$ for all $t$, the gradient step of $\OCP_z$ at time $t$ is of size $\frac{\norm{\solutionspace_{max}}}{\sqrt{t}}$. Thus, for the gradient step to be smaller than $\epsilon$, we require
$\frac{\norm{\solutionspace_{max}}}{\sqrt{t}} \leq \epsilon .$ Rearranging, and denoting the resulting task instance as $X$, we get
$$X \geq \frac{\norm{\solutionspace_{max}}^2}{\epsilon^2} .$$

Second, note that for gradient steps less than $\epsilon$, the algorithm is guaranteed to converge once the sum of gradient steps is larger than $\norm{\solutionspace_{max}}$. Formally, for convergence at task instance $Y$, we require
$$\sum^Y_{t=X} \frac{\norm{\solutionspace_{max}}}{\sqrt{t}} \geq \norm{\solutionspace_{max}}, $$

or equivalently
$$\sum^Y_{t=X} \frac{1}{\sqrt{t}} \geq 1. $$

Rewriting the left side of the inequality, we get
$$2 \sqrt{Y} - 2 \sqrt{X} \geq 1. $$

Inserting our result for $X$, and rewriting, we get
$$Y \geq \left(\frac{\norm{\solutionspace_{max}}}{\epsilon} + \frac{1}{2}\right)^2, $$
which is satisfied for the setting considered in this theorem, and thus after $B$ instances, $\w^B_{\z}$ is guaranteed to satisfy $\lvert \w^B_{\z} - c^* \rvert \leq \epsilon$.

\subsection*{Step 2}
After the end of the first batch, in the shared parameter setting weights across all tasks are equivalent after projection. Since for any solution $\w_z$ that satisfies $\lvert \w_{\z} - c^* \rvert \leq \epsilon$ the gradient is zero, after $B$ instances learners do not divert from their solution and suffer zero loss for task instances $t>B$. Thus, using Corollary \ref{corollary:COCP.special}, the loss of the \COCP algorithm in this setting is bounded by 
$$\regret_{\COCP}(T) \leq \frac{3}{2} \sqrt{B} \norm{\solutionspace_{max}}  \norm{\maxgradient_{max}}.$$

\subsection*{Step 3}
In the analysis of the \IOCP algorithm for this setting, weights of tasks are not shared and thus, after receiving $B$ task instances, every learner suffers the regret derived above. Thus, for $K$ learners, the \IOCP algorithm achieves a regret bound that is worse by up to a factor $\problems$. 
\end{proof}




\section{Tighter bound for Corollary~\ref{corollary:COCP.special}}\label{appendix_corollaries_tighter-bounds}


\begin{proof}[Proof of Corollary~\ref{corollary:COCP.special}] \leavevmode

\subsection*{Preparation}
Define $\what^t$ as in the proof of Theorem~\ref{theorem:COCP}. Note that for 
$\xi^t = 1, \dualitygap^t = 0  \ \forall t \in [T]$,
$\w^t$ is the exact projection on $\solutionspace^*$ for all $t$, and thus also $\what^t = \w^t$ for all $t$.

Our goal is to upper bound the regret 
$$ \regret_{\COCP}(T) = \sum^T_{t=1} \gradient^t_{\z^t} \cdot (\w^t_{\z^t} - \wstar_{\z^t}) .$$

We rewrite
\begin{align*}
\sum^T_{t=1} \gradient^t \cdot (\w^t - \wstar) &= \sum^T_{t=1} \gradient^t \cdot (\w^{t+1} - \wstar) + \sum^T_{t=1} \gradient^t \cdot (\w^t - \w^{t+1}) \\
&= \sum^T_{t=1} \left( \gradient^t \cdot (\w^{t+1} - \wstar) + \frac{1}{2}\gradient^t \cdot (\w^t - \w^{t+1}) \right) + \frac{1}{2} \sum^T_{t=1} \gradient^t \cdot (\w^t - \w^{t+1})
\end{align*}
 and upper bound both sums individually.

Throughout the proof we will use the Bregman divergence with the regularizer $\regularizer^t(\w) = \frac{1}{2} \w' \Q^t \w$ and apply Lemmas~\ref{lemma:update1} and~\ref{lemma:update2} to get an equivalent update procedure.

\subsection*{Step 1: First sum}
For the first part of the sum, applying Lemma~\ref{lemma:constraint-minimizer} with $\w^{t+1}$ as the constraint minimizer of the objective $\eta \gradient^t \cdot \w + \divergence_{\regularizer^t}(\w, \w^t)$ and $\wstar \in \solutionspace^*$, we have
$$ \eta \gradient^t \cdot (\w^{t+1} - \wstar) \leq \divergence_{\regularizer^t}(\wstar, \w^t) - \divergence_{\regularizer^t}(\wstar, \w^{t+1}) - \divergence_{\regularizer^t}(\w^{t+1}, \w^t) . $$

Adding over time,
\begin{align*}
\sum^T_{t=1} \eta \gradient^t \cdot (\w^{t+1} - \wstar) &\leq \sum^T_{t=1} \divergence_{\regularizer^t}(\wstar, \w^t) - \divergence_{\regularizer^t}(\wstar, \w^{t+1}) - \divergence_{\regularizer^t}(\w^{t+1}, \w^t) \\
&= \sum^T_{t=1} \left(\divergence_{\regularizer^{t+1}}(\wstar, \w^{t+1}) - \divergence_{\regularizer^t}(\wstar, \w^{t+1}) - \divergence_{\regularizer^t}(\w^{t+1}, \w^t) \right) \\
& \phantom{\leq} + \divergence_{\regularizer^1}(\wstar, \w^1) - \divergence_{\regularizer^{T+1}}(\wstar, \w^{T+1}) \\
&\leq \sum^T_{t=1} \left(\divergence_{\regularizer^{t+1}}(\wstar, \w^{t+1}) - \divergence_{\regularizer^t}(\wstar, \w^{t+1}) - \divergence_{\regularizer^t}(\w^{t+1}, \w^t) \right) \\
& \phantom{\leq} + \divergence_{\regularizer^1}(\wstar, \w^1) .
\end{align*}

We now rewrite each term on the right side of the inequality. For the first two terms, using Lemma~\ref{lemma:rtplusminusrt}, we get 
$$ \sum^T_{t=1} \divergence_{\regularizer^{t+1}}(\wstar, \w^{t+1}) - \divergence_{\regularizer^t}(\wstar, \w^{t+1}) \leq \frac{1}{2} \norm{\solutionspace_{max}}^2 \sqrt{T \problems} .$$

For the third term, we start by applying Lemma~\ref{lemma:constraint-minimizer} with $\w^{t+1}$ as the constraint minimizer of the objective $\eta \gradient^t \cdot \w + \divergence_{\regularizer^t}(\w, \w^t)$ and $\w^t \in \solutionspace^*$, such that
$$ \eta \gradient^t \cdot (\w^{t+1} - \w^t) \leq - 2 \divergence_{\regularizer^t}(\w^{t+1}, \w^t) $$
and thus

$$ \divergence_{\regularizer^t}(\w^{t+1}, \w^t) \leq \frac{1}{2} \eta \gradient^t \cdot (\w^t - \what^{t+1}) .$$

To make the inequality an equality we subtract $\da \geq 0$ from the right side, such that
$$ \divergence_{\regularizer^t}(\w^{t+1}, \w^t) = \frac{1}{2} \eta \gradient^t \cdot (\w^t - \w^{t+1}) - \da .$$

For the fourth term, using that $\Q^1$ is 1 in exactly one position, we have
$$\divergence_{\regularizer^1}(\wstar, \w^1) \leq \frac{1}{2} \norm{\solutionspace_{max}}^2 .$$ 

Dividing by $\eta$ and adding the second part of the sum, we get
$$\sum^T_{t=1} \gradient^t \cdot (\w^{t+1} - \wstar) + \frac{1}{2}\gradient^t \cdot (\w^t - \w^{t+1}) \leq \frac{1}{2 \eta} \norm{\solutionspace_{max}}^2 \sqrt{T \problems}  + \frac{1}{2 \eta} \norm{\solutionspace_{max}}^2 + \sum^T_{t=1}\frac{\da}{\eta} .$$

\subsection*{Step 2: Second sum}

To bound $\frac{1}{2} \sum^T_{t=1} \gradient^t \cdot (\w^t - \w^{t+1}) $ we start by using H{\"o}lder's inequality (see for example \cite{beckenbach2012inequalities}) to get
$$\gradient^t \cdot (\w^t - \w^{t+1})  \leq \norm{\gradient^t}^*_{\Q^t} \norm{\w^t - \w^{t+1}}_{\Q^t}$$

subtracting $\db \geq 0$ from the right side to maintain equality,
$$\gradient^t \cdot (\w^t - \w^{t+1})  = \norm{\gradient^t}^*_{\Q^t} \norm{\w^t - \w^{t+1}}_{\Q^t} - \db $$
and

$$\frac{1}{2} \gradient^t \cdot (\w^t - \w^{t+1})  = \frac{1}{2} \norm{\gradient^t}^*_{\Q^t} \norm{\w^t - \w^{t+1}}_{\Q^t} - \frac{1}{2} \db .$$

We again use
$$ \divergence_{\regularizer^t}(\w^{t+1}, \w^t) = \frac{1}{2} \eta \gradient^t \cdot (\w^t - \w^{t+1}) - \da.$$

Note that $\divergence_{\regularizer^t}(\w^t, \w^{t+1}) = \frac{1}{2} \norm{\w^t - \w^{t+1}}^2_{\Q^t}$. Thus,
$$  \frac{1}{2} \norm{\w^t - \w^{t+1}}^2_{\Q^t} = \frac{1}{2} \eta \gradient^t \cdot (\w^t - \w^{t+1}) - \da.$$
$$ \norm{\w^t - \w^{t+1}}^2_{\Q^t} = \eta \gradient^t \cdot (\w^t - \w^{t+1}) - 2 \da.$$

Using H{\"o}lder's inequality and again subtracting $\db \geq 0$ to maintain equality,
$$\norm{\w^t - \w^{t+1}}^2_{\Q^t} = \eta \norm{\gradient^t}^*_{\Q^t} \norm{\w^t - \w^{t+1}}_{\Q^t} - 2 \da - \eta \db $$

where
$$ \norm{\gradient^t}^*_{\Q^t} = \max_{\pmb{x}} \pmb{x} \cdot \gradient^t : \norm{\pmb{x}}_{\Q^t} \leq 1 .$$

Dividing by $\norm{\w^t - \w^{t+1}}_{\Q^t} $ we get
$$\norm{\w^t - \w^{t+1}}_{\Q^t} = \eta \norm{\gradient^t}^*_{\Q^t} - 2 \frac{\da}{\norm{\w^t - \w^{t+1}}_{\Q^t}} - \eta \frac{\db}{\norm{\w^t - \w^{t+1}}_{\Q^t}} .$$
Therefore,
\begin{align*}
\frac{1}{2} \gradient^t \cdot (\w^t - \w^{t+1}) &= \frac{1}{2} \norm{\gradient^t}^*_{\Q^t} \norm{\w^t - \w^{t+1}}_{\Q^t} - \frac{\db}{2} \\
&= \frac{1}{2} \eta (\norm{\gradient^t}^*_{\Q^t})^2 - \da \frac{\norm{\gradient^t}^*_{\Q^t}}{\norm{\w^t - \w^{t+1}}_{\Q^t}} - \frac{1}{2} \eta \db \frac{\norm{\gradient^t}^*_{\Q^t}}{\norm{\w^t - \w^{t+1}}_{\Q^t}} - \frac{\db}{2} .
\end{align*}

For the first term, similar to step 4 in the proof of Theorem~\ref{theorem:COCP}, summing and using
$$\norm{\gradient^t}^*_{\Q^t} \leq \norm{\gradient^t_\z}_2  \frac{1}{\sqrt{\counter^t_\z}^{1/2}}$$
we get
\begin{align*}
\sum^T_{t=1}  \frac{1}{2} \eta (\norm{\gradient^t}^*_{\Q^t})^2 &\leq \frac{1}{2} \eta \sum^T_{t=1}  \sum^\problems_{\z=1} \norm{\gradient^t_\z}^2_2 \frac{1}{\sqrt{\counter^t_\z}} \\
&\leq \frac{1}{2} \eta \sum^\problems_{\z=1}  \sum^T_{t=1} \norm{\maxgradient_\z}^2 \frac{1}{\sqrt{\counter^t_\z}}  \\
&\leq \eta \sum^\problems_{\z=1} \norm{\maxgradient_\z}^2  \left( \sqrt{\counter^T_\z} -1 \right) \\
&\leq \eta \norm{\maxgradient_{max}}^2 \sum^\problems_{\z=1} \left( \sqrt{\counter^T_\z} -1 \right) \\
&\leq \eta \norm{\maxgradient_{max}}^2 \sqrt{T\problems} - \eta \norm{\maxgradient_{max}}^2 \problems
\end{align*}

Thus,
\begin{align*}
\sum^T_{t=1} \frac{1}{2} \gradient^t \cdot (\w^t - \w^{t+1}) &\leq \eta \norm{\maxgradient_{max}}^2 \sqrt{T\problems} - \eta \norm{\maxgradient_{max}}^2 \problems \\
&\phantom{\leq} - \sum^T_{t=1} \left( \da \frac{\norm{\gradient^t}^*_{\Q^t}}{\norm{\w^t - \w^{t+1}}_{\Q^t}} - \frac{1}{2} \eta \db \frac{\norm{\gradient^t}^*_{\Q^t}}{\norm{\w^t - \w^{t+1}}_{\Q^t}} - \frac{\db}{2} \right).
\end{align*}

\subsection*{Step 3: Combination}

We first show that the sum of terms involving $\da$ and $\db$ is non-positive and can thus be upper bounded by $0$. Note that $$ \da = \frac{1}{2} \eta \gradient^t \cdot (\w^t - \w^{t+1}) - \frac{1}{2} \norm{\w^t - \w^{t+1}}^2_{\Q^t}$$ and $$\db = \norm{\gradient^t}^*_{\Q^t} \norm{\w^t - \w^{t+1}}_{\Q^t} - \gradient^t \cdot (\w^t - \w^{t+1}) .$$ 

Inserting and canceling identical terms,
\begin{align*}
&  \frac{\da}{\eta} - \da \frac{\norm{\gradient^t}^*_{\Q^t}}{\norm{\w^t - \w^{t+1}}_{\Q^t}} - \frac{1}{2} \eta \db \frac{\norm{\gradient^t}^*_{\Q^t}}{\norm{\w^t - \w^{t+1}}_{\Q^t}} - \frac{\db}{2} \\
&= \frac{1}{2} \gradient^t \cdot (\w^t - \w^{t+1}) - \frac{1}{2 \eta} \norm{\w^t - \w^{t+1}}^2_{\Q^t} \\ 
&\phantom{=} -  \frac{1}{2} \eta (\norm{\gradient^t}^*_{\Q^t})^2 + \frac{1}{2} \eta \gradient^t \cdot (\w^t - \w^{t+1}) \frac{\norm{\gradient^t}^*_{\Q^t}}{\norm{\w^t - \w^{t+1}}_{\Q^t}} \\
&\phantom{=} - \frac{1}{2} \eta \gradient^t \cdot (\w^t - \w^{t+1}) \frac{\norm{\gradient^t}^*_{\Q^t}}{\norm{\w^t - \w^{t+1}}_{\Q^t}} + \frac{1}{2} \norm{\gradient^t}^*_{\Q^t} \norm{\w^t - \w^{t+1}}_{\Q^t} \\
&\phantom{=} - \frac{1}{2} \norm{\gradient^t}^*_{\Q^t} \norm{\w^t + \w^{t+1}}_{\Q^t} + \frac{1}{2} \gradient^t \cdot (\w^t - \w^{t+1}) \\
&= \gradient^t \cdot (\w^t - \w^{t+1}) - \frac{1}{2} \eta (\norm{\gradient^t}^*_{\Q^t})^2 - \frac{1}{2 \eta} \norm{\w^t - \w^{t+1}}^2_{\Q^t} \\
&= \frac{1}{2 \eta} \left( 2 \eta \gradient^t \cdot (\w^t - \w^{t+1}) - \eta^2 (\norm{\gradient^t}^*_{\Q^t})^2 - \norm{\w^t - \w^{t+1}}^2_{\Q^t} \right) \\
&\leq - \frac{1}{2 \eta} \left( \eta \norm{\gradient^t}^*_{\Q^t} - \norm{\w^t - \w^{t+1}}^2_{\Q^t} \right)^2 ,
\end{align*}

where we used H{\"o}lder's inequality in the last step to get
$$\eta \gradient^t \cdot (\w^t - \w^{t+1})  \leq \norm{\gradient^t}^*_{\Q^t}\norm{\w^t - \w^{t+1}}^2_{\Q^t} .$$

Inserting the remaining terms, we get
$$\sum^T_{t=1} \gradient^t \cdot (\w^t - \wstar) \leq \frac{1}{2 \eta} \norm{\solutionspace_{max}}^2 \sqrt{T \problems} + \eta \norm{\maxgradient_{max}}^2 \sqrt{T\problems}  + \frac{1}{2 \eta} \norm{\solutionspace_{max}}^2 - \eta \norm{\maxgradient_{max}}^2 \problems .$$

Using $\eta = \frac{\norm{\solutionspace_{max}}}{\norm{\maxgradient_{max}}}$ and $\problems \geq 1$,
$$\sum^T_{t=1} \gradient^t \cdot (\w^t - \wstar) \leq \frac{3}{2} \sqrt{T \problems} \norm{\solutionspace_{max}} \norm{\maxgradient_{max}} .$$


\end{proof}



\section{Details of the Survey Study} \label{appendix:experiments}
We recruited workers from the MTurk platform \cite{mturk} to participate in the survey study. 
After several introductory questions about their preferences and familiarity with travel accommodations, participants were presented two randomly chosen apartments from \names{Airbnb}, using data from \names{insideairbnb.com} \cite{insideairbnb}. To choose between the apartment, participants were given the price, location, picture, number of reviews and rating of each apartment, as shown in Figure~\ref{figure:airbnbsetup}.
\vspace{1em}
\begin{figure*}[h!]
\centering
\includegraphics[width=1\textwidth]{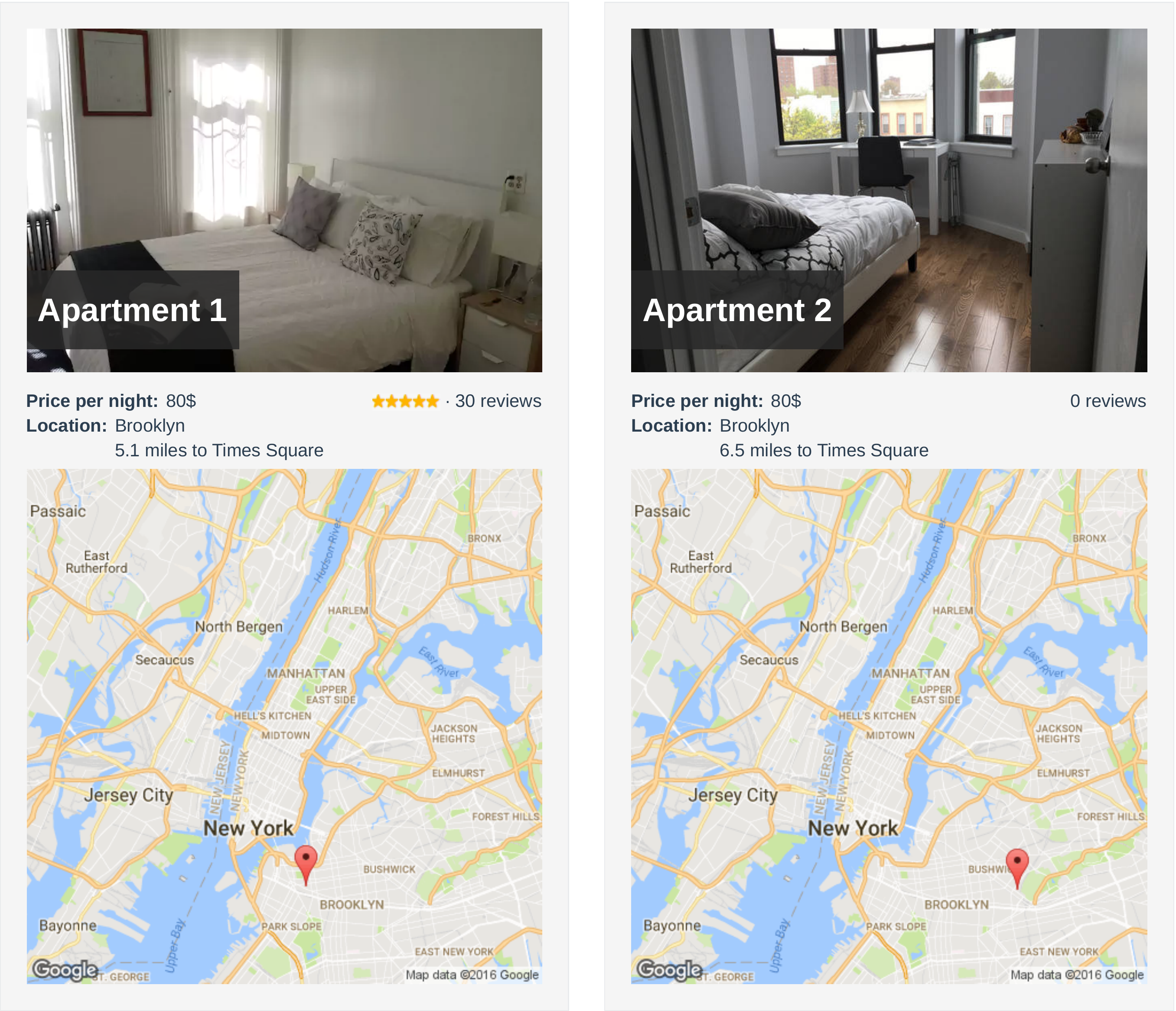} 
\vspace{2mm}
\caption{Snapshot of the survey shown to participants from MTurk.} \label{figure:airbnbsetup}
\end{figure*}

After the participants decided on their preference between the two randomly chosen apartments, they were told that the rental site would like to offer a special discount for the other apartment, which would reduce the price per night of that apartment. They were then asked to select the discount per night that they would like to receive to choose this apartment instead of their initial choice. The options for the answer of this questions were 0, 10, 20, 30, 40, more than 40, and NA, where participants were asked to select NA if they were not willing to consider the offer for any price. In total, we got $943$ responses, which are summarized in Figure~\ref{fig.discounts} in Section~\ref{sec.userstudy}.

}




\end{document}